%% file: main.tex
\documentclass[sigconf]{acmart}
\acmDOI{} 
\acmISBN{} 
\setcopyright{none} 
\AtBeginDocument{%
  }

\acmConference[KDD '23]{Proceedings of the 29th ACM SIGKDD Conference on Knowledge Discovery and Data Mining}{August 6--10, 2023}{Long Beach, CA, USA}

\usepackage{subcaption}

\usepackage{multirow}
\usepackage{diagbox}
\usepackage{graphicx}
\usepackage{booktabs}
\usepackage{hyperref}
\usepackage{url}
\usepackage[normalem]{ulem}
\usepackage{enumitem}
\usepackage{makecell}
\usepackage{soul}
\usepackage{footmisc}
\usepackage{xcolor}
\input{math_commands.tex}

\begin{document}

\title{TSMixer: Lightweight MLP-Mixer Model for \\ Multivariate Time Series Forecasting}


\author{Vijay Ekambaram}
\email{vijaye12@in.ibm.com}
\orcid{0009-0004-5824-9826}
\affiliation{
  \institution{IBM Research}
  \city{Bangalore}
  \country{India}
}

\author{Arindam Jati}
\authornote{Both authors contributed equally to this research.}
\email{arindam.jati@ibm.com}
\orcid{0000-0002-9498-8536}
\affiliation{%
  \institution{IBM Research}
  \city{Bangalore}
  \country{India}
}

\author{Nam Nguyen}
\email{nnguyen@us.ibm.com}
\orcid{0009-0008-6981-4463}
\authornotemark[1]
\affiliation{
  \institution{IBM Research}
  \city{Yorktown Heights, NY}
  \country{USA}
}

\author{Phanwadee Sinthong}
\email{gift.sinthong@ibm.com}
\orcid{0009-0006-4423-3860}
\affiliation{
  \institution{IBM Research}
  \city{Yorktown Heights, NY}
  \country{USA}
}

\author{Jayant Kalagnanam}
\email{jayant@us.ibm.com}
\orcid{009-0009-5051-2606}
\affiliation{
  \institution{IBM Research}
  \city{Yorktown Heights, NY}
  \country{USA}
}

\renewcommand{\shortauthors}{Vijay Ekambaram, Arindam Jati, Nam Nguyen, Phanwadee Sinthong, \& Jayant Kalagnanam}

\newcommand{\vijay}[1]{\textcolor{green}{Vijay: #1}}
\newcommand{\multilinecomment}[1]{}

\newcommand{\nam}[1]{\textcolor{blue}{Nam: #1}}

\newcommand{\arindam}[1]{\textcolor{red}{Arindam: #1}}

\newcommand{\gift}[1]{\textcolor{orange}{Gift: #1}}

\newcommand{\ie}{i.e., }
\newcommand{\eg}{e.g., }
\newcommand{\etc}{\textit{etc.}}
\newcommand{\wrt}{\textit{w.r.t.}~}
\newcommand{\viz}{\textit{viz. }}
\newcommand{\etal}{\textit{et. al.}}
\newcommand{\mixer}{MLP-Mixer~}
\newcommand{\tsm}{TSMixer}
\newcommand{\citsm}{CI-TSMixer}
\newcommand{\ictsm}{IC-TSMixer}
\newcommand{\vtsm}{V-TSMixer}

\newcommand{\van}{V}
\newcommand{\ci}{CI}
\newcommand{\ic}{IC}
\newcommand{\ga}{G}
\newcommand{\hr}{H}
\newcommand{\cc}{CC}
\newcommand{\tsmgh}{CI-\tsm(\ga,\hr)}
\newcommand{\tsmghc}{CI-\tsm(\ga,\hr,\cc)}
\newcommand{\tsmb}{CI-\tsm-Best}
\newcommand{\cn}{|}

\begin{abstract}
Transformers have gained popularity in time series forecasting for their ability to capture long-sequence interactions. However, their memory and compute-intensive requirements pose a critical bottleneck for long-term forecasting, despite numerous advancements in compute-aware self-attention modules. To address this, we propose \textbf{TSMixer}, a lightweight neural architecture exclusively composed of multi-layer perceptron (MLP) modules. TSMixer is designed for multivariate forecasting and representation learning on patched time series, providing an efficient alternative to Transformers.
Our model draws inspiration from the success of MLP-Mixer models in computer vision. We demonstrate the challenges involved in adapting Vision MLP-Mixer for time series and introduce empirically validated components to enhance accuracy. This includes a novel design paradigm of attaching online reconciliation heads to the MLP-Mixer backbone, for explicitly modeling the time-series properties such as hierarchy and channel-correlations. We also propose a Hybrid channel modeling approach to effectively handle noisy channel interactions and generalization across diverse datasets, a common challenge in existing patch channel-mixing methods. Additionally, a simple gated attention mechanism is introduced in the backbone to prioritize important features. By incorporating these lightweight components, we significantly enhance the learning capability of simple MLP structures, outperforming complex Transformer models with minimal computing usage. Moreover, TSMixer's modular design enables compatibility with both supervised and masked self-supervised learning methods, making it a promising building block for time-series Foundation Models.
TSMixer outperforms state-of-the-art MLP and Transformer models in forecasting by a considerable margin of 8-60\%. It also outperforms the latest strong benchmarks of Patch-Transformer models (by 1-2\%) with a significant reduction in memory and runtime (2-3X). \textcolor{blue}{The source code of our model is officially released as PatchTSMixer in the HuggingFace}\textcolor{purple}{\href{https://huggingface.co/docs/transformers/main/en/model_doc/patchtsmixer}{ [Model]}  
\href{https://github.com/ibm/tsfm\#notebooks-links}{[Examples]}}.
 \newline
\textit{\footnotesize Accepted in the Proceedings of the 29th ACM SIGKDD Conference on Knowledge Discovery and Data Mining (KDD '23), Research Track. Delayed release in arXiv to comply with the conference policies on the double-blind review process. This work has been submitted to the KDD peer-review process on Feb, 02, 2023~\cite{kdd23}.}

\end{abstract}

\maketitle

\section{Introduction}




Multivariate time series forecasting is the task of predicting the future values of multiple (possibly) related time series at future time points given the historical values of those time series.
It has widespread applications in weather forecasting, traffic prediction, industrial process controls, \textit{etc.}
This decade-long problem has been well studied in the past by several statistical and ML methods~\cite{al2018short},~\cite{hyndman_book}.
Recently, Transformer~\cite{transformer}-based models are becoming popular for long-term multivariate forecasting due to their powerful capability to capture long-sequence dependencies.
Several Transformer architectures were suitably designed for this task in last few years including Informer~\cite{informer}, Autoformer~\cite{autoformer}, FEDformer~\cite{autoformer}, and Pyraformer~\cite{pyraformer}.
However, the success of the Transformer in the semantically rich NLP domain has not been well-transferred to the time series domain.
One of the possible reasons is that, though positional embedding in Transformers preserves some ordering information, the nature of the permutation-invariant self-attention mechanism inevitably results in temporal information loss. This hypothesis has been empirically validated in~\cite{dlinear}, where an embarrassingly simple linear~(DLinear) model is able to outperform most of the above-mentioned Transformer-based forecasting models. 

Furthermore, unlike words in a sentence, individual time points in a time series lack significant semantic information and can be easily inferred from neighboring points. Consequently, a considerable amount of modeling capacity is wasted on learning point-wise details. PatchTST~\cite{patchtst} has addressed this issue by dividing the input time series into patches and applying a transformer model, resulting in superior performance compared to existing models.
However, PatchTST employs a pure channel\footnote{Throughout the text, we will use ``channel'' to denote the individual time series in a multivariate data (i.e. a multivariate time series is a multi-channel signal).} independence approach which does not explicitly capture the cross-channel correlations. 
PatchTST has demonstrated that channel independence can enhance performance compared to channel mixing, where channels are simply concatenated before being fed into the model. This simple mixing approach can lead to noisy interactions between channels in the initial layer of the Transformer, making it challenging to disentangle them at the output. CrossFormer~\cite{crossformer}, another recent effort on Patch Transformers with an improved channel mixing technique, also faces this issue (see Appendix Table 9). Therefore, there is a necessity to explicitly model channel interactions to seize opportunistic accuracy improvements while effectively reducing the high volume of noisy interactions across channels. In addition, though PatchTST reduces the timing and memory overhead via patching, it still uses the multi-head self-attention under the hood, which is computationally expensive even when it is applied at the patch level.

Recently, a series of multi-layer perceptron~(MLP) models under the umbrella of ``MLP-Mixers" have been proposed in the computer vision domain~\cite{mlpmixer, payattention, resmlp}.
These models are lightweight and fast, and achieve comparable or superior performance to vision Transformer models while completely eliminating the need for computing intense multi-head self-attention~\cite{mlpmixer}. Moreover, MLP-Mixer by its default architecture does not disturb the temporal ordering of the inputs 
which makes it a natural fit for time-series problems and resolves the concerns raised in DLinear~\cite{dlinear}.
At this point, we ask the following important question - \textit{Can MLP-Mixer models yield superior performance for multivariate time series forecasting?  If so, what are the required time series customizations?}

Towards this, we show that the adoption of \mixer for time series is \textit{not} trivial, \ie just applying the vanilla \mixer with some input and output shape modification will not make it a powerful model, and will have suboptimal performance compared to the state-of-the-arts. Hence, we propose \textbf{~\tsm}, a novel MLP-Mixer architecture for accurate multivariate time series forecasting. Like PatchTST, \tsm~ is also patching-based and follows a modular architecture of learning a common \textit{``backbone''} to capture the temporal dynamics of the data as a patch representation, and different \textit{``heads''} are attached and finetuned based on various downstream tasks (Ex. forecasting).  Backbone is considered task-independent and can learn across multiple datasets with a masked reconstruction loss while the heads are task and data-specific. \\

\textbf{Key highlights of ~\tsm~}are as follows: 
\begin{enumerate}
    \item TSMixer is a patching-based, lightweight neural architecture designed solely with MLP modules that exploits various inherent time-series characteristics for accurate \textit{multivariate forecasting and representation learning}.
    \item TSMixer proposes a novel design paradigm of attaching \& tuning online reconciliation\footnote{\label{reconcile_tag}This ``reconciliation'' is different from the standard reconciliation in hierarchical forecasting~\cite{hyndman_book}. Here, reconciliation targets patch-aggregation and  cross-channel correlation and it is done online during training. We call it ``online" as it is learned as part of the overall loss computation and not as a separate offline process.} heads to the MLP-Mixer backbone that significantly empowers the learning capability of simple MLP structures to outperform complex Transformer models while using less computing resources. This study is the first of its kind to highlight the benefits of infusing online reconciliation approaches in the prediction head of Mixer style backbone architectures for time-series modeling. 
    \item Specifically, ~\tsm~ proposes two novel online reconciliation heads to tune \& improve the forecasts by leveraging the following intrinsic time series properties:  \textit{hierarchical patch-aggregation} and \textit{cross-channel correlation}.
    \item  For effective \textit{cross-channel modeling}, ~\tsm~ follows a novel ``Hybrid" channel modeling by \textit{augmenting a channel independent backbone with a cross-channel reconciliation head}. This hybrid architecture allows the backbone to generalize across diverse datasets with different channels, while the reconciliation head effectively learns the channel interactions specific to the task and data. This approach effectively handles noisy interactions across channels, which is a challenge in existing patch channel-mixing methods.   
    \item For effective \textit{long sequence modeling}, ~\tsm~ introduces a simple Gated Attention\footnote{\label{ga_note}Gated Attention proposed in this paper is different from the Spatial Gating Unit proposed in gMLP~\cite{payattention}} that guides the model to focus on the important features. This, when augmented with Hierarchical Patch Reconciliation Head and standard MLP-Mixer operations enables effective modeling of long-sequence interaction and eliminates the need for complex multi-head self-attention blocks.
    \item Finally, the modular design of ~\tsm~ enables it to work with both supervised and masked self-supervised learning methodologies which makes it a potential building block for time-series foundation models~\cite{foundation}.
\end{enumerate}
We conduct a detailed empirical analysis on 7 popular public datasets, wherein. ~\tsm~ outperforms all existing benchmarks with \textit{extremely reduced training time and memory usage}. Snapshot (relative MSE improvements) of the primary benchmarks are as follows:
\begin{itemize}
    \item \tsm~ outperforms DLinear by a considerable margin of 8\% (Table~\ref{tab:supervised}).
    \item \tsm~ marginally outperforms PatchTST by 1-2\% but with a significant 2-3X reduction in training time and memory usage. (Table~\ref{tab:supervised},\ref{tab:speedup},\ref{tab:fm_1},\ref{tab:fm-2-all}).
    \item \tsm~ outperforms all other state-of-the-arts by a significant margin of 20-60\%. 
\end{itemize}

\section{Related Work}
\textbf{Transformer-based time series models: }
The success of Transformers~\cite{transformer} in the NLP domain has inspired time series researchers to come up with TS Transformer-based models.
Due to the quadratic time and memory complexity of Transformer architectures, one major difficulty in adopting NLP Transformers for time series tasks lies in processing much longer input sequence lengths ($L$). 
To address this, several compute-aware time series Transformer models such as Informer~\cite{informer}, Autoformer~\cite{autoformer}, Pyraformer\cite{pyraformer} and FEDformer~\cite{fedformer} have been proposed, which modifies the self-attention mechanisms to achieve $O(L)$ or $O(L\log(L))$ complexity.


\textbf{Patch-based time series models:}
The above-mentioned works mostly feed granular-level time series into the Transformer model, hence, they focus on learning attention over every time point. In contrast, PatchTST~\cite{patchtst} and CrossFormer~\cite{crossformer} enable patching before feeding it to the Transformers for learning representation across patches. PatchTST follows the channel independence approach and in contrast, CrossFormer follows the channel-mixing approach.

\textbf{\mixer in vision domain:}
Recently, multi-layer perceptron's (MLP) strength has been reinvented in the vision domain~\cite{mlpmixer, payattention, resmlp}.
The idea of~\cite{mlpmixer} is to transform the input image through a series of permutation operations, which inherently enforces mixing of features within and across patches to capture short and long-term dependencies without the need for self-attention blocks.
\mixer attains similar performance as CNN and Transformer. To extend upon, in gMLP~\cite{payattention}, authors propose to use MLP-Mixer with Spatial Gating Unit (SGU)\footref{ga_note} and ResMLP~\cite{resmlp} proposes to use MLP-Mixer with residual connections. 

\textbf{MLP based sequence / time series models:}\footnote{Comparison with \cite{chen2023} is not included, as it was released in arXiv one month after the start of the KDD 2023 peer-review process~\cite{kdd23}, where our current paper has been submitted and got accepted.}
Similar efforts have recently been put in the time series domain.
Zhang \etal~\cite{lightts} proposed LightTS that is built upon MLP and two sophisticated downsampling strategies for improved forecasting.
The DLinear model~\cite{dlinear} also employs a simple linear model after decomposing the time series into trend and seasonality components. DLinear questions the effectiveness of Transformers as it easily beats Transformer-based SOTAs.
Li \etal~\cite{MLP4Rec} proposed MLP4Rec, a pure MLP-based architecture for sequential recommendation task, to capture correlations across time, channel, and feature dimensions.


\section{Methodology}
\begin{figure}
    \centering
    \includegraphics[width=0.8\columnwidth]{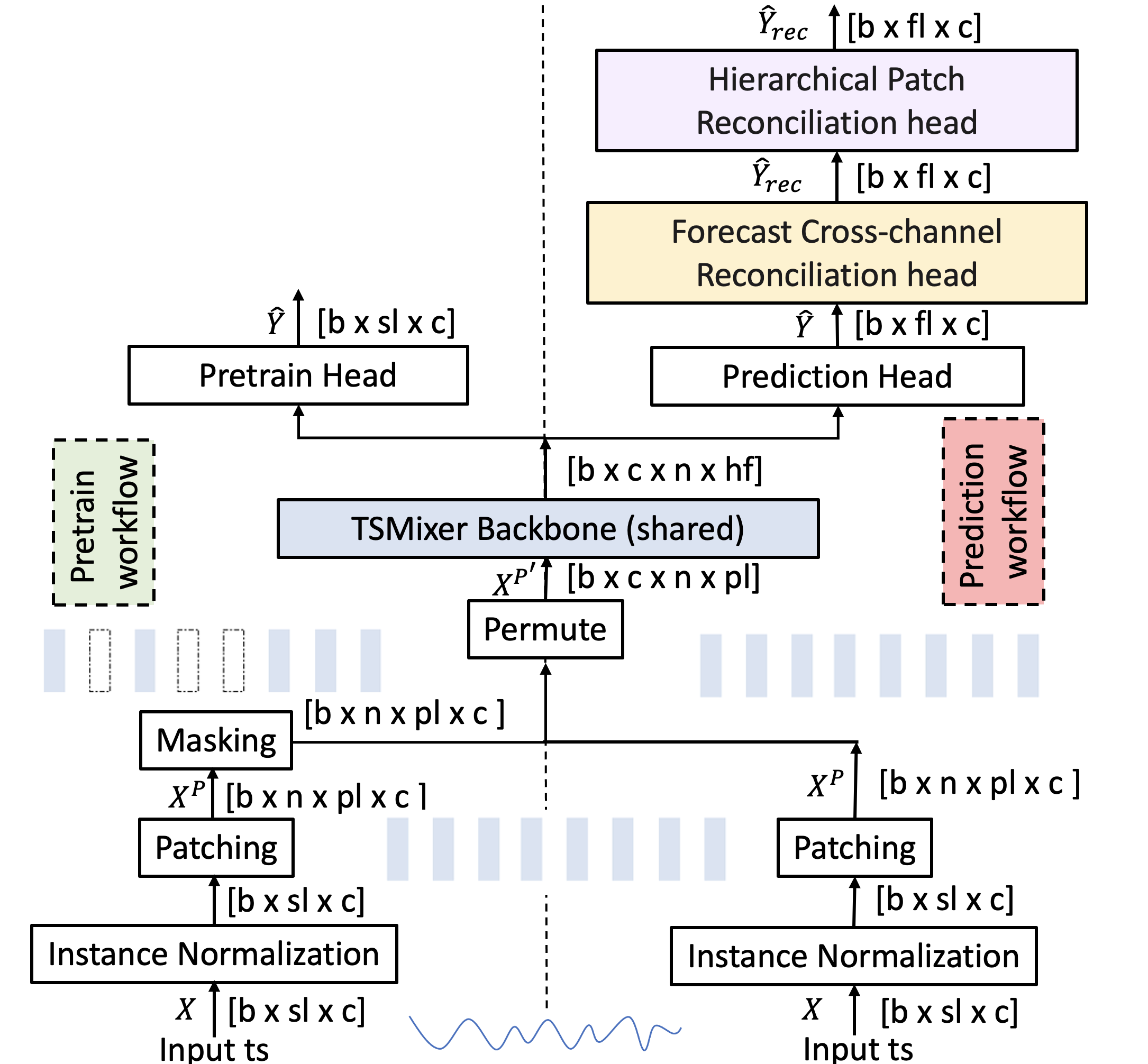}
    \caption{High-level model architecture.}
    \label{fig:high-level}
\end{figure}

\subsection{Notations}
Throughout the paper, we use the following variable names: 
$\mX_{c \times L}$: a multivariate time series of length $L$ and $c$ channels or $c$ time series, $sl \le L$: input sequence length, $fl$: forecast sequence length (a.k.a. horizon), 
$b$: batch size, $n$: number of patches, $pl$: patch length, $hf$: hidden feature dimension, $ef$: expansion feature dimension, $nl$: number of \mixer~ layers, $\gM$: learned DNN model, $op$: number of output forecast patches, $cl$: context length, $spl$: patch length for cross-channel forecast reconciliation head, $\hat{H}$: patch-aggregated prediction, $H$: patch-aggregated ground truth, $\hat{Y}_\text{rec}$: actual base prediction, $Y_\text{rec}$: base ground truth, $sf$: scale factor.
To ensure better understanding, we provide the shape of a tensor as its subscript in the text, and in square brackets in the architectural diagrams.
We denote a linear layer in the neural network by $\gA(\cdot)$ for compactness. 

The multivariate forecasting task is defined as predicting future values of the time series given some history:
\begin{equation}
    \hat{\mY}_{fl \times c} = \gM \left( \mX_{sl \times c} \right).
\end{equation}
The ground truth future values will be denoted by ${\mY}_{fl \times c}$.

\begin{figure*}[t]
     \centering
     \begin{subfigure}[b]{0.85\columnwidth}
         \centering
         \includegraphics[width=\columnwidth]{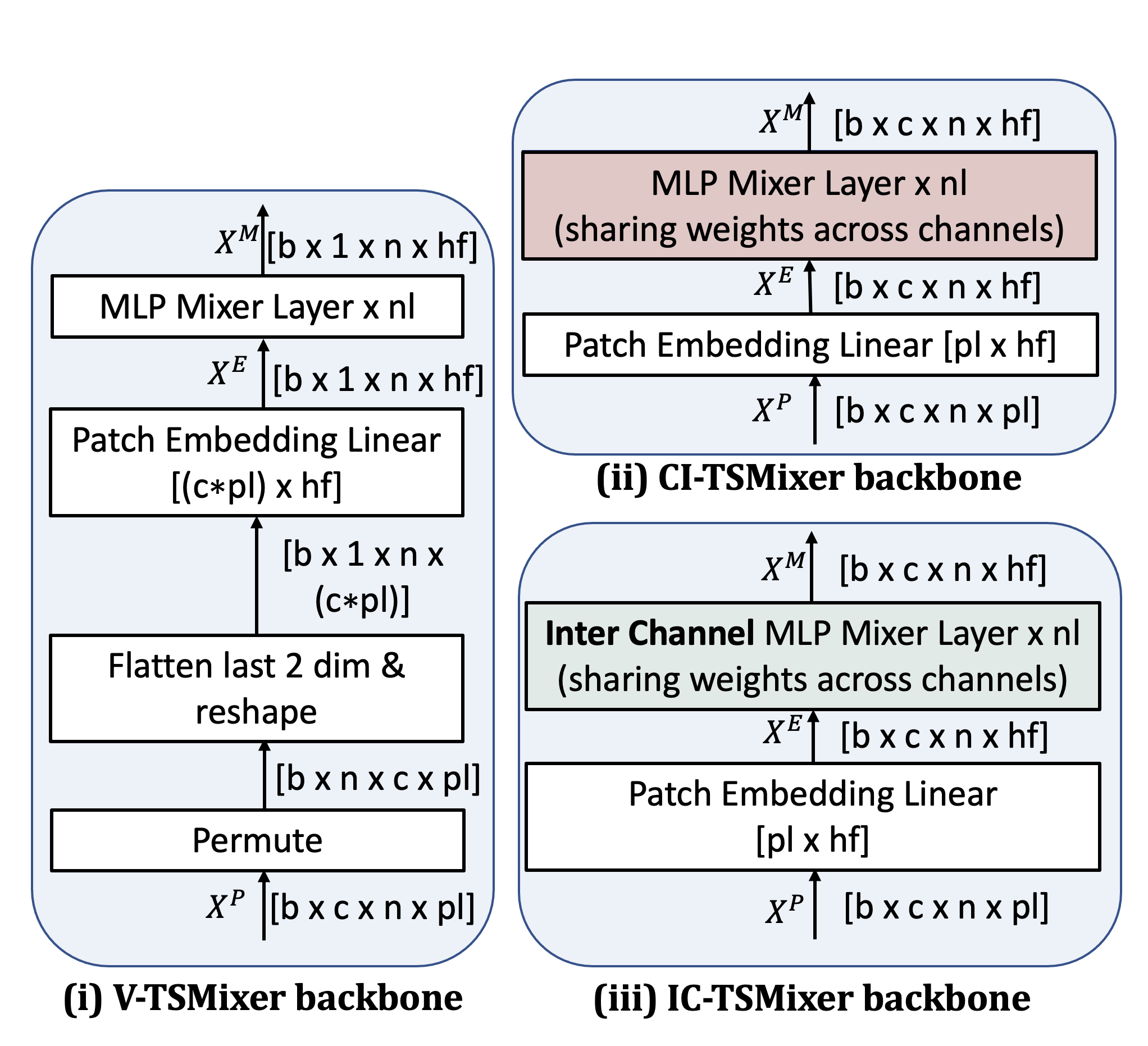}
         \caption{Different backbones in TSMixer}
         \label{fig:backbone:backbones}
     \end{subfigure}
     \hfill
     \begin{subfigure}[b]{1.1\columnwidth}
         \centering
         \includegraphics[width=\columnwidth]{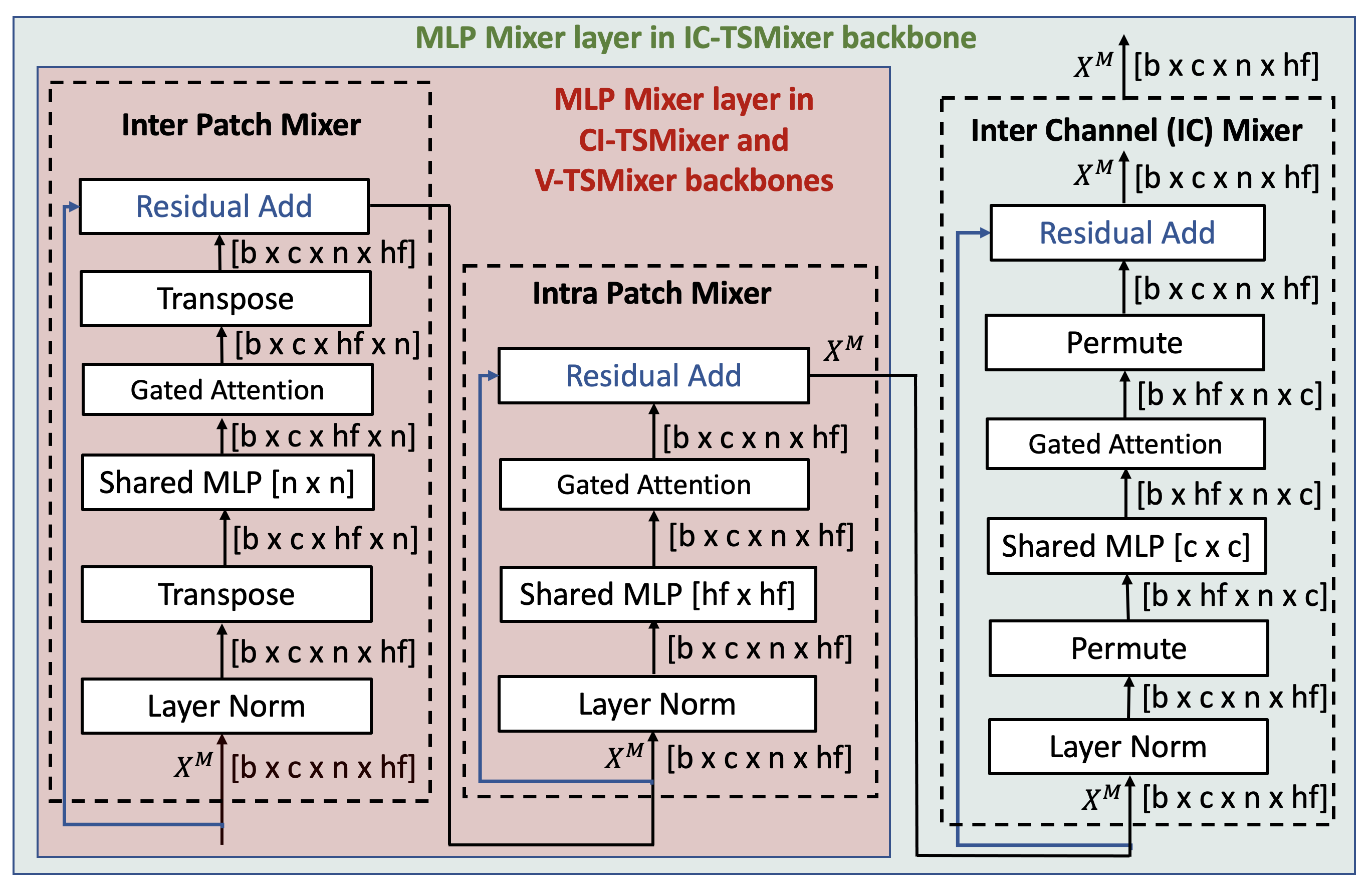}
         \caption{MLP Mixer layer architecture}
         \label{fig:backbone:mixer-layers}
     \end{subfigure}
    \caption{
    Different backbones in TSMixer and the architecture of the mixer layers.
    }
    \label{fig:backbone}
\end{figure*}

\subsection{Training methodologies}\label{subsec:Training methodologies}
TSMixer has two training methodologies: supervised and self supervised.
The supervised training follows the ``prediction'' workflow as shown in the right part of Figure~\ref{fig:high-level}.
First, the input history time series goes through a sequence of transformation (normalization, patching, and permutation).
Then, it enters the TSMixer backbone which is responsible for the main learning process.
The prediction head converts the output embeddings of the backbone to the base forecasts, $\hat{\mY}$.
The model can be trained to minimize the mean squared error~(MSE) of the base forecasts: 
$\gL\left( \mY, \hat{\mY} \right) = \left|\left| \mY - \hat{\mY} \right|\right|_2^2$.
We introduce two extra \textit{online} forecast reconciliation heads which, if activated, can tune the base forecasts and produce more accurate forecasts by leveraging cross-channel and patch-aggregation information. When any or both of these reconciliation heads are activated, a customized MSE-based objective function is employed on the tuned forecasts. A detailed discussion is provided in Section~\ref{subsubsec:Forecast online reconciliation}. 

The self-supervised training is performed in two stages.
First, the model is pretrained (see the ``pretrain'' workflow in Figure~\ref{fig:high-level}) with a self-supervised objective. Then, the pretrained model is finetuned through the ``prediction'' workflow for a supervised downstream task.
Self-supervised pretraining has been found to be useful for a variety of NLP~\cite{bert}, vision~\cite{beit}, and time series tasks~\cite{patchtst}.
Similar to BERT's~\cite{bert} masked language modeling~(MLM) in the NLP domain, we employ a masked time series modeling~(MTSM) task.
The MTSM task randomly applies masks on a fraction of input patches, and the model is trained to recover the masked patches from the unmasked input patches.
Other input transformations in the pretrain workflow are the same as in the prediction workflow.
The MTSM task minimizes the MSE reconstruction error on the masked patches. The modular design of TSMixer enables it for either supervised or self-supervised training by only changing the model head (and keeping backbone the same).

\subsection{Model components} 
Here we discuss the modeling components that are introduced to a vanilla \mixer to have improved performance.
The high-level architecture is shown in Figure~\ref{fig:high-level}.
For stochastic gradient descent~(SGD), each minibatch, $\mX^{B}_{b \times sl \times c}$, is populated from $\mX$ by a moving window technique.
The forward pass of a minibatch along with its shape is shown in Figure~\ref{fig:high-level}

\subsubsection{\textbf{Instance normalization}}
The input time series segment goes through reversible instance normalization~(RevIN)~\cite{revin}.
RevIN standardizes the data distribution (\ie removes mean and divides by the standard deviation) to tackle data shifts in the time series.

\subsubsection{\textbf{Patching}}
\label{sec:patching}
Every univariate time series is segregated into overlapping / non-overlapping patches with a stride of $s$. For self-supervised training flow, patches have to be strictly non-overlapping.
The minibatch $\mX^B_{b \times sl \times c}$ is reshaped into $\mX^P_{b \times n \times pl \times c}$, where $pl$ denotes the patch length, and $n$ is the number of patches (hence, $n = \lfloor (sl - pl)/s \rfloor +1$).
The patched data is then permuted to $\mX^{P'}_{b \times c \times n \times pl}$ and fed to the TSMixer backbone model.
Patching reduces the number of input tokens to the model by a factor of $s$, and hence, increases the model runtime performance significantly as compared to standard point-wise Transformer approaches~\cite{patchtst}.

\subsubsection{\textbf{TSMixer backbone}}
Three possible backbones are shown in Figure~\ref{fig:backbone:backbones}.
The vanilla backbone (\vtsm) flattens the channel and patch dimensions $(c \times pl)$ on the input before passing to the next layer. This approach is commonly followed in Vision MLP Mixing techniques and hence acts as a vanilla baseline.
We propose two new types of backbone: channel independent backbone (\citsm) and inter-channel backbone (\ictsm). They differ in their MLP mixer layer architectures.
\citsm~backbone is inspired from the PatchTST~\cite{patchtst} model, in which the MLP Mixer layer is shared across channels which forces the model to share learnable weights across the channels. This results in a reduction of model parameters. Moreover, \citsm~ enables us to employ TSMixer for self-supervised modeling over multiple datasets each having a different number of channels, which ~\vtsm~ cannot.
In ~\ictsm, an extra inter-channel mixer module is activated in the backbone to explicitly capture inter-channel dependencies. All these backbones will be compared with extensive experimentation.

Note that all the TSMixer backbones start with a linear patch embedding layer (see Figure~\ref{fig:backbone:backbones}).
It transforms every patch independently into an embedding: $\mX^E_{b \times c \times n \times hf} = \gA \left( \mX^{P'} \right)$.
The weight and bias of $\gA(\cdot)$ are shared across channels for ~\citsm~ and ~\ictsm ~backbones, but \textit{not} for ~\vtsm. Since channels are completely flattened in ~\vtsm, $\gA(\cdot)$ in ~\vtsm~ does not have any notion of multiple channels (i.e. c = 1).

\subsubsection{\textbf{MLP Mixer layers}}
\tsm~ backbone stacks a set of mixer layers like encoder stacking in Transformers. Intuitively, each mixer layer (Figure~\ref{fig:backbone:mixer-layers}) tries to learn correlations across three different directions: (i) between different patches, (ii) between the hidden feature inside a patch, and (iii) between different channels.
The former two mixing methods are adopted from the vision \mixer, while the last one is proposed particularly for multivariate time series data.
The \textbf{inter patch mixer} module employs a shared MLP (weight dimension $=n\times n$) to learn correlation between different patches. The \textbf{intra patch mixer} block's shared MLP layer mixes the dimensions of the hidden features, and hence the weight matrix has a dimension of $hf \times hf$. The proposed \textbf{inter channel mixer} (weight matrix size $=c \times c$) mixes the input channel dimensions, and tries to capture correlations between multiple channels in a multivariate context. This inter-channel mixer has been proposed in MLP4Rec~\cite{MLP4Rec} for the event prediction problem and we investigate its applicability for the time-series domain. Please note that inter-channel mixer block is included only in the ~\ictsm~ backbone and not with the \citsm~ and \vtsm~ backbones (Figure~\ref{fig:backbone:mixer-layers}). 
The input and output of the mixer layers and mixer blocks are denoted by $\mX^M_{b \times c \times n \times hf}$. Based on the dimension under focus in each mixer block, the input gets reshaped accordingly to learn correlation along the focussed dimension. The reshape gets reverted in the end to retain the original input shape 
across the blocks and layers. 
All of the three mixer modules are equipped with an MLP block (Appendix Figure.~\ref{fig:mlp}), layer normalization~\cite{layernorm}, residual connection~\cite{resnet}, and gated attention. The former three components are standard in \mixer while the gated attention block is described below. 


\subsubsection{\textbf{Gated attention~(GA) block}}

Time series data often has a lot of unimportant features that confuse the model. In order to effectively filter out these features, we add a simple Gated Attention ~\cite{bahdanauattention} after the MLP block in each mixer component. GA acts like a simple gating function that probabilistically upscales the dominant features and downscales the unimportant features based on its feature values. The attention weights are derived by: $\mW^A_{b \times c \times n \times hf} = \text{softmax}\left( \gA \left( \mX^M \right) \right)$.
The output of the gated attention module is obtained by performing a dot product between the attention weights and the hidden tensor coming out of the mixer modules: $\mX^G = \mW^A \cdot \mX^M$ (Appendix Figure.~\ref{fig:ga}). Augmenting GA with standard mixer operations effectively guides the model to focus on the important features leading to improved long-term interaction modeling, without requiring the need for complex multi-head self-attention.


\subsubsection{\textbf{Model heads}}

Based on the training methodology (i.e. supervised or self-supervised), we either add prediction or pretrain head to the backbone. Both heads employ a simple Linear layer with dropout after flattening the hidden features across all patches (Appendix Figure.~\ref{fig:model-heads}). By default, heads share the same weights across channels. The output of the prediction head is the predicted multivariate time series ($\hat{\mY}_{ b \times fl \times c }$), while the pretrain head emits a multivariate series of the same dimension as the input ($\hat{\mX}_{ b \times sl \times c}$).


\subsubsection{\textbf{Forecast online reconciliation}}\label{subsubsec:Forecast online reconciliation}
Here we propose two novel methods (in the prediction workflow, see Figure~\ref{fig:high-level}) to tune the original forecasts, $\hat{\mY}$, based on two important characteristics of time series data: inherent temporal hierarchical structure and cross-channel dependency. Any or both of them can be activated in our TSMixer model to obtain reconciled forecasts.

\textbf{Cross-channel forecast reconciliation head:}
\begin{figure}
    \centering
    \includegraphics[width=0.85\columnwidth]{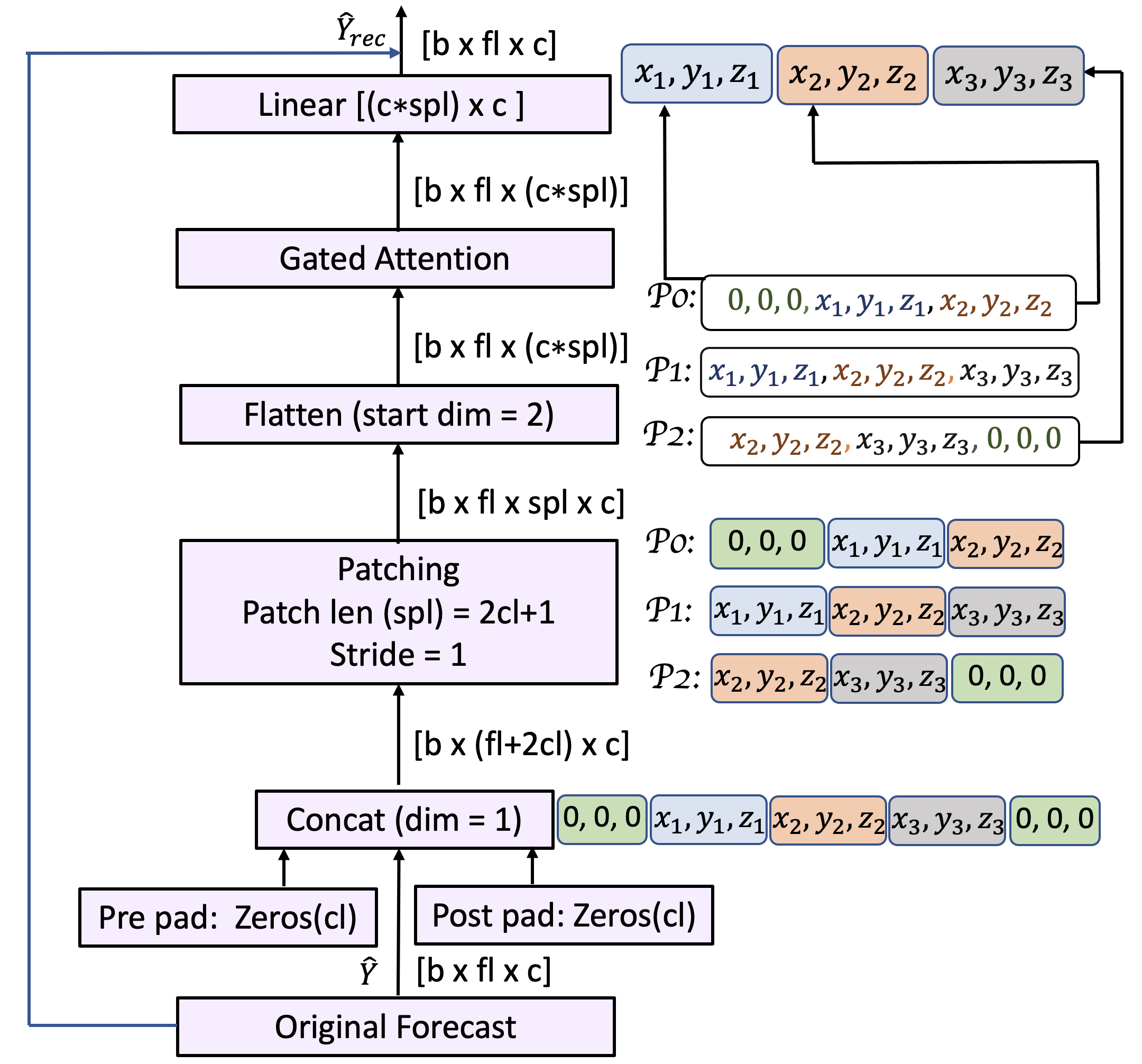}
    \caption{Cross-channel forecast reconciliation head.}
    \label{fig:FCR}
\end{figure}
In many scenarios, the forecast for a channel at a certain time might be dependent on the forecast of another channel at a different point in time on the horizon.
For example, in retail domain, sales at a future time point might be dependent on the discount patterns around that time.
Hence, we introduce a cross-channel forecast reconciliation head which derives an objective that attempts to learn cross-dependency across channels within a local surrounding context \textit{in the forecast horizon}.
Figure~\ref{fig:FCR} demonstrates its architecture.

First, every forecast point is converted into a patch (of length $spl$) by appending its pre and post-surrounding forecasts based on the context length ($cl$). 
Then, each patch is flattened across channels and passed through a gated attention and linear layer to obtain a revised forecast point for that patch. 
Thus, all channels of a forecast point reconcile its values based on the forecast channel values in the surrounding context leading to effective cross-channel modeling. Residual connections ensure that reconciliation does not lead to accuracy drops in scenarios when the channel correlations are very noisy.
Since the revised forecasts have the same dimension as the original forecasts, no change to the loss function is required. From experiments, we observe that the ``hybrid" approach of having a channel-independent backbone augmented with a cross-channel reconciliation head provides stable improvements as compared to other channel-mixing approaches. Also, this architecture helps in better backbone generalization as it can be trained with multiple datasets with a varying number of channels while offloading the channel-correlation modeling to the prediction head (which is task and data-dependent).

\textbf{Online hierarchical patch reconciliation head:}\footnote{This ``reconciliation'' is different from the reconciliation in hierarchical forecasting~\cite{hyndman_book}. Here, hierarchy is defined as an aggregation over a patch, and the reconciliation is done online during training.}.
\begin{figure}
    \centering
    \includegraphics[width=0.73\columnwidth]{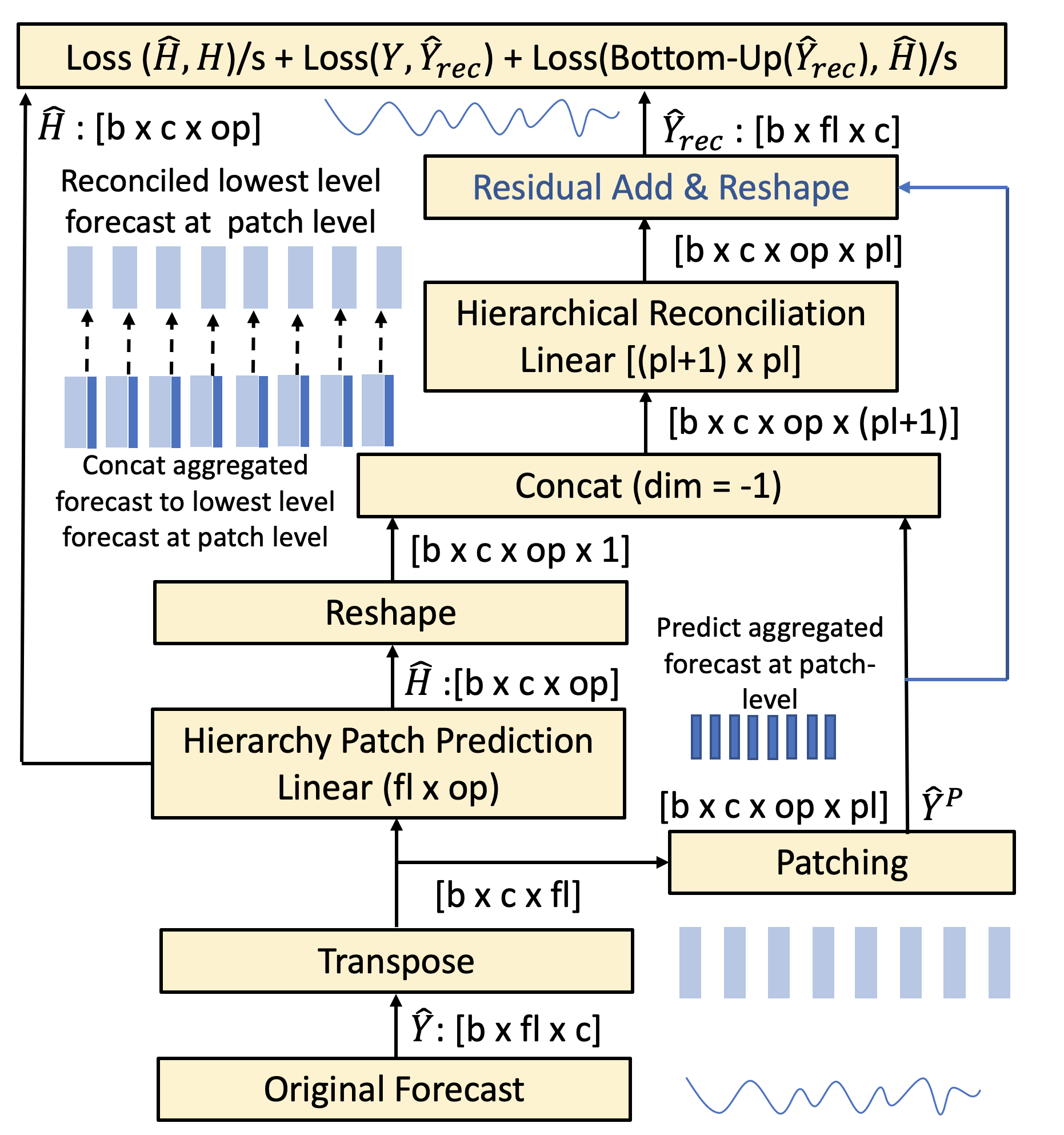}
    \caption{Online hierarchical patch reconciliation head.}
    \label{fig:HPR}
\end{figure}
Time series data often possess an inherent hierarchical structure, either explicitly known (\eg hierarchical forecasting datasets~\cite{hyndman_book}), or as an implicit characteristic (\eg an aggregation of weather forecasts for seven days denotes an estimate of weekly forecast, an aggregation of sales forecast over all stores in a state denotes state-level sales, and so on).
In general, aggregated time series have better predictability and a good forecaster aims at achieving low forecast error in all levels of the hierarchy (\cite{hyndman_book, makridakis2022m5, proxyhierarchical}).
Here, we propose a novel method to automatically derive a hierarchical patch aggregation loss (online during training) that is minimized along with the granular-level forecast error.
Figure~\ref{fig:HPR} shows the architecture.
The original forecast $\hat{\mY}$ is segregated into $op$ number of patches, each of length $pl$.
We denote this as $\hat{\mY}^P$.
Now, $\hat{\mY}$ is also passed through a linear layer to predict the hierarchically aggregated forecasts at the patch-level: $\hat{\mH}_{b \times c \times op}= \gA \left( \hat{\mY}_{b \times c \times fl} \right)$.
Then, we concatenate $\hat{\mY}^P$ and $\hat{\mH}$ at the patch level and pass it through another linear transformation to obtain reconciled granular-level forecast: $\hat{\mY}_{\text{rec}}$. Thus, the granular-level forecasts get reconciled at a patch level based on the patch-aggregated forecasts leading to improved granular-level forecasts. Residual connections ensure that the reconciliation does not lead to accuracy drops in scenarios when predicting aggregated signals become challenging.
Now, the hierarchical patch aggregation loss is calculated as follows:
\begin{equation}
    \gL_{\text{hier}} = \frac{1}{sf} \gL\left( \hat{\mH}, \mH \right) +
    \gL \left( \mY, \hat{\mY}_{\text{rec}} \right) +
    \frac{1}{sf} \gL \left( \text{BU} \left(  \hat{\mY}_{\text{rec}} \right), \hat{\mH} \right)
\end{equation}
where, $\mY$ is ground-truth future time series, $H$ is the aggregated ground-truth at patch-level, $\text{BU}$ refers to bottom-up aggregation of the granular-level forecasts to obtain the aggregated patch-level forecasts~\cite{hyndman_book}, and $sf$ is scale factor.
For MSE loss, $sf = (pl)^2$.
More intuitively, this loss tries to tune the base forecasts in a way such that they are not only accurate at the granular-level, but also accurate at the aggregated patch-level.
Note that a pre-defined dataset-specific hierarchical structure can be enforced here, but it is left for future studies.

\begin{table}[htbp]
\centering
\setlength{\tabcolsep}{2.5pt}
\scalebox{0.85}{
\begin{tabular}{c|ccccccc}
\toprule  
Datasets & Weather  & Traffic & Electricity & ETTH1 & ETTH2 & ETTM1 & ETTM2 \\
\midrule  
Features & 21 & 862 & 321 & 7 & 7 & 7 & 7 \\
Timesteps & 52696 & 17544 & 26304 & 17420 & 17420 & 69680 & 69680 \\
\bottomrule 
\end{tabular}}
\caption{Statistics of popular datasets for benchmark.}
\label{table:dataset}
\end{table}

\begin{table*}[htb]
	\centering
        \scalebox{0.9}{
		  \begin{tabular}{cc|c|cc|cc|cc|cc|cc|ccc}
            \cline{2-15}
			&\multicolumn{2}{c|}{Models} & \multicolumn{2}{c}{\textbf{\citsm-Best}} & \multicolumn{2}{c|}{DLinear} & \multicolumn{2}{c|}{PatchTST}& \multicolumn{2}{c|}{FEDformer}& \multicolumn{2}{c|}{Autoformer}& \multicolumn{2}{c}{Informer} \\
			\cline{2-15}
			&\multicolumn{2}{c|}{Metric}&MSE&MAE&MSE&MAE&MSE&MAE&MSE&MAE&MSE&MAE&MSE&MAE\\
			\cline{2-15}
            &\multirow{4}*{\rotatebox{90}{ETTH1}} & 96 & \textbf{0.368$\pm$0.001} & \textbf{0.398$\pm$0.001} & 0.375 & \uline{0.399} & \uline{0.370} & 0.400 & 0.376 & 0.419 & 0.449 & 0.459 & 0.865 & 0.713 \\ 
            &\multicolumn{1}{c|}{} & 192 & \textbf{0.399$\pm$0.002} & \uline{0.418$\pm$0.001} & \uline{0.405} & \textbf{0.416} & 0.413 & 0.429 & 0.420 & 0.448 & 0.500 & 0.482 & 1.008 & 0.792 \\ 
            &\multicolumn{1}{c|}{} & 336 & \textbf{0.421$\pm$0.004} & \textbf{0.436$\pm$0.003} & 0.439 & 0.443 & \uline{0.422} & \uline{0.440} & 0.459 & 0.465 & 0.521 & 0.496 & 1.107 & 0.809 \\ 
            &\multicolumn{1}{c|}{} & 720 & \textbf{0.444$\pm$0.003} & \textbf{0.467$\pm$0.002} & 0.472 & 0.490 & \uline{0.447} & \uline{0.468} & 0.506 & 0.507 & 0.514 & 0.512 & 1.181 & 0.865 \\ 
            \cline{2-15} 
            &\multirow{4}*{\rotatebox{90}{ETTH2}} & 96 & \uline{0.276$\pm$0.006} & \textbf{0.337$\pm$0.003} & 0.289 & \uline{0.353} & \textbf{0.274} & \textbf{0.337} & 0.346 & 0.388 & 0.358 & 0.397 & 3.755 & 1.525 \\ 
            &\multicolumn{1}{c|}{} & 192 & \textbf{0.330$\pm$0.003} & \textbf{0.374$\pm$0.001} & 0.383 & 0.418 & \uline{0.341} & \uline{0.382} & 0.429 & 0.439 & 0.456 & 0.452 & 5.602 & 1.931 \\ 
            &\multicolumn{1}{c|}{} & 336 & \uline{0.357$\pm$0.001} & \uline{0.401$\pm$0.002} & 0.448 & 0.465 & \textbf{0.329} & \textbf{0.384} & 0.496 & 0.487 & 0.482 & 0.486 & 4.721 & 1.835 \\ 
            &\multicolumn{1}{c|}{} & 720 & \uline{0.395$\pm$0.003} & \uline{0.436$\pm$0.003} & 0.605 & 0.551 & \textbf{0.379} & \textbf{0.422} & 0.463 & 0.474 & 0.515 & 0.511 & 3.647 & 1.625 \\ 
            \cline{2-15} 
            &\multirow{4}*{\rotatebox{90}{ETTM1}} & 96 & \textbf{0.291$\pm$0.002} & \uline{0.346$\pm$0.002} & 0.299 & \textbf{0.343} & \uline{0.293} & \uline{0.346} & 0.379 & 0.419 & 0.505 & 0.475 & 0.672 & 0.571 \\ 
            &\multicolumn{1}{c|}{} & 192 & \textbf{0.333$\pm$0.002} & \uline{0.369$\pm$0.002} & \uline{0.335} & \textbf{0.365} & \textbf{0.333} & 0.370 & 0.426 & 0.441 & 0.553 & 0.496 & 0.795 & 0.669 \\ 
            &\multicolumn{1}{c|}{} & 336 & \textbf{0.365$\pm$0.005} & \textbf{0.385$\pm$0.004} & \uline{0.369} & \uline{0.386} & \uline{0.369} & 0.392 & 0.445 & 0.459 & 0.621 & 0.537 & 1.212 & 0.871 \\ 
            &\multicolumn{1}{c|}{} & 720 & \textbf{0.416$\pm$0.002} & \textbf{0.413$\pm$0.001} & \uline{0.425} & 0.421 & \textbf{0.416} & \uline{0.420} & 0.543 & 0.490 & 0.671 & 0.561 & 1.166 & 0.823 \\ 
            \cline{2-15} 
            &\multirow{4}*{\rotatebox{90}{ETTM2}} & 96 & \textbf{0.164$\pm$0.002} & \textbf{0.255$\pm$0.002} & 0.167 & 0.260 & \uline{0.166} & \uline{0.256} & 0.203 & 0.287 & 0.255 & 0.339 & 0.365 & 0.453 \\ 
            &\multicolumn{1}{c|}{} & 192 & \textbf{0.219$\pm$0.002} & \textbf{0.293$\pm$0.002} & 0.224 & 0.303 & \uline{0.223} & \uline{0.296} & 0.269 & 0.328 & 0.281 & 0.340 & 0.533 & 0.563 \\ 
            &\multicolumn{1}{c|}{} & 336 & \textbf{0.273$\pm$0.003} & \textbf{0.329$\pm$0.003} & 0.281 & \uline{0.342} & \uline{0.274} & \textbf{0.329} & 0.325 & 0.366 & 0.339 & 0.372 & 1.363 & 0.887 \\ 
            &\multicolumn{1}{c|}{} & 720 & \textbf{0.358$\pm$0.002} & \textbf{0.380$\pm$0.001} & 0.397 & 0.421 & \uline{0.362} & \uline{0.385} & 0.421 & 0.415 & 0.433 & 0.432 & 3.379 & 1.338 \\ 
            \cline{2-15} 
            &\multirow{4}*{\rotatebox{90}{Electricity}} & 96 & \textbf{0.129$\pm$1e-4} & \uline{0.224$\pm$0.001} & \uline{0.140} & 0.237 & \textbf{0.129} & \textbf{0.222} & 0.193 & 0.308 & 0.201 & 0.317 & 0.274 & 0.368 \\ 
            &\multicolumn{1}{c|}{} & 192 & \textbf{0.146$\pm$0.001} & \uline{0.242$\pm$1e-4} & 0.153 & 0.249 & \uline{0.147} & \textbf{0.240} & 0.201 & 0.315 & 0.222 & 0.334 & 0.296 & 0.386 \\ 
            &\multicolumn{1}{c|}{} & 336 & \textbf{0.158$\pm$0.001} & \textbf{0.256$\pm$0.001} & 0.169 & 0.267 & \uline{0.163} & \uline{0.259} & 0.214 & 0.329 & 0.231 & 0.338 & 0.300 & 0.394 \\ 
            &\multicolumn{1}{c|}{} & 720 & \textbf{0.186$\pm$0.001} & \textbf{0.282$\pm$0.001} & 0.203 & 0.301 & \uline{0.197} & \uline{0.290} & 0.246 & 0.355 & 0.254 & 0.361 & 0.373 & 0.439 \\ 
            \cline{2-15} 
            &\multirow{4}*{\rotatebox{90}{Traffic}} & 96 & \textbf{0.356$\pm$0.002} & \textbf{0.248$\pm$0.002} & 0.410 & 0.282 & \uline{0.360} & \uline{0.249} & 0.587 & 0.366 & 0.613 & 0.388 & 0.719 & 0.391 \\ 
            &\multicolumn{1}{c|}{} & 192 & \textbf{0.377$\pm$0.003} & \uline{0.257$\pm$0.002} & 0.423 & 0.287 & \uline{0.379} & \textbf{0.256} & 0.604 & 0.373 & 0.616 & 0.382 & 0.696 & 0.379 \\ 
            &\multicolumn{1}{c|}{} & 336 & \textbf{0.385$\pm$0.002} & \textbf{0.262$\pm$0.001} & 0.436 & 0.296 & \uline{0.392} & \uline{0.264} & 0.621 & 0.383 & 0.622 & 0.337 & 0.777 & 0.420 \\ 
            &\multicolumn{1}{c|}{} & 720 & \textbf{0.424$\pm$0.001} & \textbf{0.283$\pm$0.001} & 0.466 & 0.315 & \uline{0.432} & \uline{0.286} & 0.626 & 0.382 & 0.660 & 0.408 & 0.864 & 0.472 \\ 
            \cline{2-15} 
            &\multirow{4}*{\rotatebox{90}{Weather}} & 96 & \textbf{0.146$\pm$0.001} & \textbf{0.197$\pm$0.002} & 0.176 & 0.237 & \uline{0.149} & \uline{0.198} & 0.217 & 0.296 & 0.266 & 0.336 & 0.300 & 0.384 \\ 
            &\multicolumn{1}{c|}{} & 192 & \textbf{0.191$\pm$0.001} & \textbf{0.240$\pm$0.001} & 0.220 & 0.282 & \uline{0.194} & \uline{0.241} & 0.276 & 0.336 & 0.307 & 0.367 & 0.598 & 0.544 \\ 
            &\multicolumn{1}{c|}{} & 336 & \textbf{0.243$\pm$0.001} & \textbf{0.279$\pm$0.002} & 0.265 & 0.319 & \uline{0.245} & \uline{0.282} & 0.339 & 0.380 & 0.359 & 0.395 & 0.578 & 0.523 \\ 
            &\multicolumn{1}{c|}{} & 720 & \uline{0.316$\pm$0.001} & \textbf{0.333$\pm$0.002} & 0.323 & 0.362 & \textbf{0.314} & \uline{0.334} & 0.403 & 0.428 & 0.419 & 0.428 & 1.059 & 0.741 \\ 
            \cline{2-15} 
            &\multicolumn{4}{c|}{\makecell{\textbf{\citsm-Best} \textbf{\% improvement (MSE)}}}& \multicolumn{2}{c}{\textbf{8\%}} & \multicolumn{2}{c}{ \textbf{1\%}}  & \multicolumn{2}{c}{\textbf{23\%}}  & \multicolumn{2}{c}{\textbf{30\%}}  & \multicolumn{2}{c}{\textbf{64\%}}  \\
            \cline{2-15} 
		\end{tabular}
	}
	\caption{Comparing TSMixer with popular benchmarks in supervised long-term multivariate forecasting. The best results are in bold and the second best is underlined. PatchTST results are reported from ~\cite{patchtst}. All other benchmarks are reported from ~\cite{dlinear}}
	\label{tab:supervised}
\end{table*}

\section{Experiments}

\subsection{Experimental settings}

\subsubsection{\textbf{Datasets}}
We evaluate the performance of the proposed \tsm~model on 7 popular multivariate datasets as depicted in Table~\ref{table:dataset}. These datasets have been extensively used in the literature~\cite{patchtst}\cite{dlinear}\cite{autoformer} for benchmarking multivariate forecasting models and are publically available in ~\citep{dlinearrepo}. We follow the same data loading parameters (Ex. train/val/test split ratio) as followed in ~\cite{patchtst}.

\subsubsection{\textbf{Model Variants}}

A ~\tsm~ variant is represented using the naming convention ${``BackboneType"}$-\tsm(${``Enhancements"}$).\\
${``BackboneType"}$ can either be Vanilla (\van), or Channel Independent (\ci), or Inter Channel (\ic).
${``Enhancements"}$ can be a combination of Gated Attention (\ga), Hierarchical Patch Reconciliation head (\hr
), and/or Cross-channel Reconciliation head (\cc). Common variants are:
\begin{itemize}
    \item \textbf{~\ci-\tsm}: Channel Independent ~\tsm~ with no enhancements
    \item \textbf{~\tsmgh}: Channel Independent ~\tsm~ with Gated Attention and Hierarchy Reconciliation head
    \item \textbf{~\tsmghc}: Channel Independent ~\tsm~ with Gated Attention, Hierarchy and Cross-Channel Reconciliation head
    \item \textbf{~\tsmb}: Best of top performing ~\tsm~variants [i.e. ~\tsmgh~ and ~\tsmghc]
    \item \textbf{\vtsm}: Vanilla ~\tsm
    \item \textbf{\ictsm}: Inter-Channel ~\tsm
\end{itemize}
Other model variants can be formed using the same naming convention. Unless stated explicitly, the cross-channel reconciliation head uses the default context length of 1.

\subsubsection{\textbf{Data \& Model Configuration}}

By default, the following data and model configuration is used: Input Sequence length $sl$ = 512, Patch length $pl$ = 16, Stride $s$ = 8, Batch size $b$ = 8, Forecast sequence length $fl \in \{96, 192, 336, 720\}$, Number of Mixer layers $nl$ = 8, feature scaler $fs$ = 2, Hidden feature size $hf$ = $fs*pl$ (32), Expansion feature size $ef$ = $fs*hf$ (64) and Dropout $do$ = 0.1. Training is performed in a distributed fashion with 8 GPUs, 10 CPUs and 1000 GB of memory. For ETT datasets, we use a lower hardware and model configuration with high dropout to avoid over-fitting, as the dataset is relatively small ($nl$ = 3, $do$ = 0.7, $ngpus$ = 1). Supervised training is performed with 100 epochs. 
In self-supervised training, we first pretrain the backbone with 100 epochs. After that, in the finetuning phase, we freeze the backbone weights for the first 20 epochs to train/bootstrap the head (also known as \textit{linear probing}), and then, we finetune the entire network (backbone + head) for the next 100 epochs. 
We choose the final model based on the best validation score. Since overlapping patches have to be avoided in self-supervised methodology, we use reduced patch length and stride with the same size there (i.e. $pl$ = 8, $s$ = 8). This further updates the hidden feature and expansion feature size by $fs$ (i.e. $hf$ = 16, $ef$ = 32 ) for self-supervised methodology.
Every experiment is executed with 5 random seeds (from 42-46) and the mean scores are reported. Standard deviation is also reported for the primary results. We use mean squared error (MSE) and mean absolute error (MAE) as the standard error metrics.


\subsubsection{\textbf{SOTA Benchmarks}}

We categorize SOTA forecasting benchmarks into the following categories: \textbf{ (i) Standard Transformers:} FEDformer~\cite{fedformer}, Autoformer\cite{autoformer} and Informer ~\cite{informer}, \textbf{(ii) Patch Transformers:} PatchTST~\cite{patchtst} and CrossFormer~\cite{crossformer}, \textbf{(iii) MLPs and Non-Transformers:} DLinear~\cite{dlinear}, LightTS~\cite{lightts} and S4~\cite{s4}, \textbf{ (iv) Self-supervised models:} BTSF~\cite{btsf}, TNC~\cite{tnc}, TS-TCC~\cite{ts-tcc}, CPC~\cite{cpc}, and TS2Vec~\cite{ts2vec}.



\subsection{Supervised Multivariate Forecasting}
In this section, we compare the accuracy and computational improvements of TSMixer with the popular benchmarks in supervised multivariate forecasting.

\subsubsection{\textbf{Accuracy Improvements}}
In Table~\ref{tab:supervised}, we compare the accuracy of \tsm~ Best variant (i.e. ~\tsmb) with SOTA benchmarks.
Since we observe similar relative patterns in MSE and MAE, we explain all the results in this paper using the MSE metric. 
~\tsm~ outperforms standard Transformer and MLP benchmarks by a significant margin (DLinear: 8\%, FEDformer:  23\%, Autoformer: 30\% and Informer: 64\%). 
PatchTST (refers to PatchTST/64 in ~\cite{patchtst}) is one of the strongest baselines, and \tsm~ marginally outperforms it by 1\%. 
However, \tsm~ achieves this improvement over PatchTST with a significant performance improvement \wrt training time and memory (Section.~\ref{sec:speedup}). 
For exhaustive results with the individual best~\tsm~ variants, refer to Appendix Table~\ref{tab:supervised-ab1}. 
Also, Appendix~\ref{appendix_1} highlights the superior performance of TSMixer with other secondary benchmarks (LightTS, S4 and CrossFormer).

\subsubsection{\textbf{Computational Improvements}}
\label{sec:speedup}
PatchTST is considered the most compute-effective model across all time series Transformers, as patching drastically reduces the input size by a factor of s (stride)~\cite{patchtst}. In contrast, TSMixer not only enables patching but also completely eliminates the self-attention blocks.  Hence, TSMixer shows significant computation improvement over PatchTST and other Transformer models.  In  Table~\ref{tab:speedup}, we highlight the speed-up and memory comparison of ~\tsm~ over PatchTST. For analysis, we capture the following metrics: (i) Multiply-Add cumulative operations on the entire data per epoch (MACs), (ii) Number of model parameters (NPARAMS), (iii) Single EPOCH TIME and (iv) Peak GPU memory reached during a training run (MAX MEMORY). For this experiment, we trained TSMixer and PatchTST models in a single GPU node with the same hardware configuration in a non-distributed manner to report the results. To ensure fairness in comparison, we use the exact model parameters of PatchTST and ~\tsm~ which were used in the error metric comparison reported in Table~\ref{tab:supervised}. In Table~\ref{tab:speedup}, we capture the average improvement of ~\tsm~ over PatchTST for each performance metric across the three larger datasets (Electricity, Weather, Traffic) with $fl$=96. Since ~\citsm~ is purely MLP based, it significantly reduces the average MACs (by $\sim$4X), NPARAMS \& MAX MEMORY (by $\sim$3X), and EPOCH TIME \footnote{MACs and EPOCH TIME are highly correlated and in general create a similar relative impact across models. However, since PatchTST involves high parallelism across the attention heads, we observe a different relative impact \wrt MACs and EPOCH TIME.} (by $\sim$2X).  Even after enabling gated attention and hierarchy reconciliation, ~\tsmgh~ still shows a good reduction in MACs \& NPARAMS (by $\sim$3X), and EPOCH TIME \& MAX MEMORY (by $\sim$2X). It is important to note that, when cross-channel reconciliation is enabled [i.e ~\tsmghc], the number of parameters becomes very high in ~\tsm~ as compared to PatchTST. The reason is that the number of parameters in the cross-channel reconciliation head is dependent on the number of channels in the dataset that leads to this scaling effect. For example, since Electricity and Traffic datasets have a very high number of channels (i.e. 321 and 862 respectively), the number of parameters of the model also scales up, whereas weather (with only 21 channels) did not encounter any such scaling effect. Even PatchTST should show a similar scaling effect if the channel correlation is enabled in it. However, even with increased parameters, ~\tsmghc~ still shows a notable reduction in MACs (by $\sim$3X) and EPOCH TIME \& MAX MEMORY (by $\sim$2X).  The reason is that parameter scaling affects only the reconciliation head and not the backbone which primarily constitutes the total training time and memory. Thus, ~\tsm~ and its variants can easily produce improved results over PatchTST in significantly less training time and memory utilization.
\begin{table}
\centering
\setlength{\tabcolsep}{1.5pt}
\scalebox{0.9}{
\begin{tabular}{c|c|c|c|c|c|c|c|c}
\hline  
Metric & Data & \makecell{Patch\\TST} & \multicolumn{2}{c|}{\makecell{\citsm\\(Avg. Imp)}} & \multicolumn{2}{c|}{\makecell{\citsm\\(\ga,\hr)\\(Avg. Imp)}} & \multicolumn{2}{c}{\makecell{\citsm\\(\ga,\hr,\cc)\\(Avg. Imp)}}  \\
\hline
\multirow{3}{*}{\makecell{MACs\\(T)}}  & Electricity & 147.879 & 37.062 &\multirow{3}*{\rotatebox{90}{\textbf{(4X)}}} & 46.44 &\multirow{3}*{\rotatebox{90}{\textbf{(3.2X)}}} & 48.566 &\multirow{3}*{\rotatebox{90}{\textbf{(3X)}}} \\ 
& Traffic & 260.367 & 65.25 &\multicolumn{1}{c|}{} & 81.77 &\multicolumn{1}{c|}{} & 91.78 &\multicolumn{1}{c}{} \\ 
& Weather & 19.709 & 4.94 &\multicolumn{1}{c|}{} & 6.19 &\multicolumn{1}{c|}{} & 6.21 &\multicolumn{1}{c}{} \\ \hline

\multirow{3}{*}{\makecell{NPARAMS \\ (M)}}   & Electricity & 1.174 & 0.348 &\multirow{3}*{\rotatebox{90}{\textbf{(3.4X)}}} & 0.4 &\multirow{3}*{\rotatebox{90}{\textbf{(2.9X)}}} & 1.648 &\multirow{3}*{\rotatebox{90}{\textbf{(1.2X)}}} \\ 
 & Traffic & 1.174 & 0.348 &\multicolumn{1}{c|}{} & 0.4 &\multicolumn{1}{c|}{} & 9.33 &\multicolumn{1}{c}{} \\ 
 & Weather & 1.174 & 0.348 &\multicolumn{1}{c|}{} & 0.4 &\multicolumn{1}{c|}{} & 0.405 &\multicolumn{1}{c}{} \\ \hline

\multirow{3}{*}{\makecell{EPOCH \\ TIME \\(min)}}  & Electricity & 36.22 & 15.56  &\multirow{3}*{\rotatebox{90}{\textbf{(2X)}}} & 20.43 &\multirow{3}*{\rotatebox{90}{\textbf{(1.6X)}}} & 20.5 &\multirow{3}*{\rotatebox{90}{\textbf{(1.5X)}}} \\ 
 & Traffic & 64.04 & 27.49 &\multicolumn{1}{c|}{} & 36.08 &\multicolumn{1}{c|}{} & 36.51 &\multicolumn{1}{c}{} \\ 
 & Weather & 5.2 & 3.08 &\multicolumn{1}{c|}{} & 4.13 &\multicolumn{1}{c|}{} & 4.17 &\multicolumn{1}{c}{} \\ \hline
 
\multirow{3}{*}{\makecell{MAX  \\ MEMORY \\ (GB)}}  & Electricity & 6.14 & 2.25 &\multirow{3}*{\rotatebox{90}{\textbf{(2.7X)}}} & 2.9 &\multirow{3}*{\rotatebox{90}{\textbf{(2.1X)}}} & 2.94 &\multirow{3}*{\rotatebox{90}{\textbf{(2X)}}} \\ 
& Traffic & 8.24 & 3.03 &\multicolumn{1}{c|}{} & 3.89 &\multicolumn{1}{c|}{} & 4.15 &\multicolumn{1}{c}{} \\ 
& Weather & 0.451 & 0.165 &\multicolumn{1}{c|}{} & 0.21 &\multicolumn{1}{c|}{} & 0.211 &\multicolumn{1}{c}{} \\ \hline
\end{tabular}}
\caption{Computational Improvement. \textbf{nX} denotes \textbf{n} times average improvement across datasets (Avg. Imp).}
\label{tab:speedup}
\end{table}

\subsection{Component \& Design Choice Analysis}
In this section, we study the impact of various key components \& design choices followed in the TSMixer.  

\subsubsection{\textbf{Effect of CI, Gated Attention \& Hierarchy Patch Reconciliation}}
Table~\ref{tab:ci_ga_hr} depicts the improvements of various enhancement components in ~\tsm~ over ~\vtsm~ in three datasets: ETTH1, ETTM1, and Weather (for space constraint). 
~\vtsm~ represents the vanilla model where all channels are flattened and processed together (similar to vision \mixer). ~\citsm~ outperforms ~\vtsm~ by 9.5\% by introducing channel independence (CI) in the backbone instead of channel flattening. By further adding gated attention (G) and hierarchy reconciliation (H) together [i.e. ~\tsmgh], we observe an additional 2\% improvement leading to a total of 11.5\% improvement \wrt ~\vtsm. 
On analysis with all datasets (Appendix Table~\ref{tab:ga_hc-ab1}), we observe that ~\tsmgh~ outperforms ~\vtsm~ by an avg. of 19.3\%. 
In general, adding `G' and `H' together leads to more stable improvements in ~\citsm~ as compared to just adding `G' or `H'.

\begin{table}
\centering
\setlength{\tabcolsep}{2pt}
\scalebox{0.9}{
\begin{tabular}{c|c|c|c|c|c|c}
\hline  
& \makecell{FL} & \makecell{V-\\\tsm} & \makecell{CI-\\\tsm} & \makecell{\citsm\\(\ga)} & \makecell{\citsm\\(\hr)} & \makecell{\citsm\\(\ga,\hr)}  \\
\hline
\multirow{4}{*}{\rotatebox{90}{ETTH1}} & 96 & 0.449 & 0.375 & 0.375 & 0.377 & \textbf{0.368} \\ 
& 192 & 0.485 & 0.411 & 0.408 & 0.410 & \textbf{0.399} \\ 
& 336 & 0.504 & 0.437 & 0.433 & 0.431 & \textbf{0.421} \\ 
& 720 & 0.573 & 0.465 & 0.454 & 0.457 & \textbf{0.444} \\ \hline

\multirow{4}{*}{\rotatebox{90}{ETTM1}} & 96 & 0.328 & 0.305 & \textbf{0.296} & 0.301 & 0.297 \\ 
& 192 & 0.372 & 0.336 & 0.334 & 0.338 & \textbf{0.333} \\ 
& 336 & 0.405 & 0.375 & 0.366 & 0.376 & \textbf{0.365} \\ 
& 720 & 0.457 & 0.425 & 0.419 & 0.422 & \textbf{0.416} \\ \hline

\multirow{4}{*}{\rotatebox{90}{Weather}}  & 96 & 0.159 & 0.150 & 0.149 & 0.151 & \textbf{0.148} \\ 
& 192 & 0.207 & 0.195 & 0.195 & 0.195 & \textbf{0.193} \\ 
& 336 & 0.256 & 0.246 & 0.246 & 0.246 & \textbf{0.243} \\ 
& 720 & 0.330 & 0.323 & 0.317 & 0.321 & \textbf{0.317} \\ \hline
\multicolumn{3}{c|}{ \makecell{\textbf{\% improvement} \\ \textbf{over \vtsm}}} & \textbf{9.5\%}	& \textbf{10.5\%} & \textbf{10\%} & \textbf{11.5\%} \\ \hline
\end{tabular}}
\caption{Effect of CI, Gated Attention and Hierarchy Reconciliation over Vanilla~\tsm~(MSE).}
\label{tab:ci_ga_hr}
\end{table}

\begin{table}
\centering
\setlength{\tabcolsep}{2pt}
\scalebox{.9}{
\begin{tabular}{c|c|c|c|c|c}
\hline  
& \makecell{FL} & \makecell{V-\tsm} & \makecell{CI-\tsm} & \makecell{IC-\tsm} & \makecell{\citsm\\(\ga,\cc-Best)}  \\
\hline
\multirow{4}{*}{\rotatebox{90}{ETTH1}} & 96 & 0.449 & 0.375 & 0.379 & \textbf{0.373} \\ 
    & 192 & 0.485 & 0.411 & 0.416  & \textbf{0.407} \\ 
    & 336 & 0.504 & 0.437 & 0.437  & \textbf{0.43} \\ 
     & 720 & 0.573 & 0.465 & 0.471  & \textbf{0.454} \\  \hline
\multirow{4}{*}{\rotatebox{90}{ETTH2}} & 96 & 0.369 & 0.284 & 0.291  & \textbf{0.269} \\
     & 192 & 0.391 & 0.353 & 0.345  & \textbf{0.330} \\ 
     & 336 & 0.403 & 0.365 & 0.361  & \textbf{0.359} \\ 
     & 720 & 0.475 & 0.406 & 0.419  & \textbf{0.393 }\\  \hline
\multirow{4}{*}{\rotatebox{90}{Weather}} & 96 & 0.159 & 0.150 & 0.150  & \textbf{0.146} \\
     & 192 & 0.207 & 0.195 & 0.196  & \textbf{0.194} \\ 
     & 336 & 0.256 & \textbf{0.246} & 0.248  & \textbf{0.246} \\ 
     & 720 & 0.330 & 0.323 & 0.338 & \textbf{0.317} \\  \hline

\multicolumn{3}{c|}{ \makecell{\textbf{\% improvement} \\ \textbf{over \vtsm}}} & \textbf{11.5\%}	& \textbf{10.7\%}  & \textbf{13.5\%} \\ \hline
\end{tabular}}
\caption{Channel mixing technique comparison (MSE).}
\label{tab:channel_table}
\end{table}

\subsubsection{\textbf{Hybrid Channel Modelling Approach.}}
In Table~\ref{tab:channel_table}, we compare various channel-mixing techniques and highlight the benefits of ``hybrid'' channel modeling followed in TSMixer. In summary, \citsm~ outperforms \vtsm~ by 11.5\%, and by adding cross-channel reconciliation head (CC), the accuracy further improves by 2\% leading to an overall improvement of 13.5\% for \citsm(\ga,\cc-Best) (i.e. channel independent backbone with CC head)
Context Length ($cl$) is an important hyperparameter in the CC head, which decides the surrounding context space across channels to enable reconciliation, and this parameter has to be decided based on the underlying data characteristics. For this experiment, we varied $cl$ from 1 to 5 and selected the best which is depicted as \citsm(\ga,\cc-Best) in Table~\ref{tab:channel_table}. 
For more exhaustive results on various $cl$ and all datasets, refer to the Appendix Table.~\ref{tab:cc-detail}. 
Moreover, we observe from Table~\ref{tab:channel_table} that \citsm(\ga,\cc-Best) performs better compared to the ~\ictsm~ which applies cross-channel correlation inside the backbone. In addition, CrossFormer~\cite{crossformer} proposes an alternative patch cross-channel correlation approach which ~\tsm~ significantly outperforms by 30\% (Appendix Table.~\ref{tab:crossformer}). Thus, the ``hybrid" channel modeling approach of having a \textit{channel-independent backbone augmented with a cross-channel reconciliation head} is relatively robust to the noisy interactions across the channels.
Also, from the architectural perspective - this approach helps in pretraining the backbone on multiple datasets with a varying number of channels. This cannot be achieved trivially with channel-mixing backbones. 

\subsection{Forecasting via Representation Learning}
In this section, we compare TSMixer with popular benchmarks on multivariate forecasting via self-supervised representation learning.

\subsubsection{\textbf{Accuracy Improvements}}
In Table~\ref{tab:fm_1}, we compare the forecasting results of ~\tsm~ with self-supervised benchmarks. For BTSF, TNC, TS-TCC and CPC - we report the results from ~\cite{btsf}. For self-supervised PatchTST, we report ETTH1 from ~\cite{patchtst} and calculated it for Weather as it was unavailable. We use the $fl$ space commonly reported across~\cite{btsf}~and~\cite{patchtst}.
In the self-supervised workflow, the TSMixer backbone is first pre-trained to learn a generic patch representation, and then the entire network (backbone + head) is finetuned for the forecasting task. By training ~\tsm~ with this approach, we observe from Table.~\ref{tab:fm_1} that ~\tsmb~ achieves a significant improvement (50-70\% margin) from existing forecasting benchmarks learned via self-supervision (such as BFSF, TNC, TS-TCC, CPC). Similar trends were also observed with TS2Vec~\cite{ts2vec} (Appendix Table~\ref{tab:ab_ts2vec}). Also, ~\tsmb~ beats self-supervised PatchTST by a  margin of 2\%. However, as explained in Section.~\ref{sec:speedup}, we achieve this improvement over PatchTST with a significant reduction in time and memory. 
For a drill-down view of ~\tsmb, refer to Appendix Table.~\ref{tab:ab_fm_1}


\subsubsection{\textbf{Pretrain strategies in Self Supervision}}
In Table.~\ref{tab:fm-2-all}, we do a detailed analysis of the self-supervised approach on 3 large datasets (Electricity, Traffic, and Weather) with 3 different pretrain data creation strategies as follows: (i) SAME (same primary data for pretrain and finetune), (ii) ALL (pretrain with all data [ETT, Electricity, Traffic and Weather] and finetune with the primary data), (iii) TL (transfer learning: pretrain with all data except primary data and finetune with the primary data). 
Since we learn across multiple datasets, we use a bigger model size for this experiment (i.e. $nl$ = 12, $fs$ = 3) for better modeling capacity. From Table.~\ref{tab:fm-2-all}, we observe that all three considered data strategies work equally well with marginal variations across them.
However, the benefits of transfer learning across multiple datasets are not very notable, as we observe in other domains (like vision and text).
On average, ~\citsm-Overall-Best(SS) [which indicates the best result across the considered data strategy variants] shows improved performance \wrt self-supervised PatchTST (reported from ~\cite{patchtst}) and supervised ~\tsm~ by 1.5\%. Thus, enabling self-supervision before the forecasting task helps in improving forecast accuracy. Furthermore, it helps in faster convergence for downstream tasks. 

\begin{table}
\centering
\setlength{\tabcolsep}{2pt}
\scalebox{0.9}{
\begin{tabular}{c|c|c|c|c|c|c|c}
\hline  
& \makecell{FL} & \makecell{CI-\tsm\\-Best} & \makecell{PatchTST} & \makecell{BTSF} & \makecell{TNC} & \makecell{TS-TCC} & \makecell{CPC}  \\
\hline
\multirow{5}*{\rotatebox{90}{ETTH1}} & 24 & \textbf{0.314} & \uline{0.322} & 0.541 & 0.632 & 0.653 & 0.687 \\ 
\multicolumn{1}{c|}{} & 48 & \textbf{0.343} & \uline{0.354} & 0.613 & 0.705 & 0.720 & 0.779 \\ 
\multicolumn{1}{c|}{} & 168& \textbf{0.397} & \uline{0.419} & 0.640 & 1.097 & 1.129 & 1.282 \\ 
\multicolumn{1}{c|}{} & 336 & \textbf{0.424} & \uline{0.445} & 0.864 & 1.454 & 1.492 & 1.641 \\ 
\multicolumn{1}{c|}{} & 720 & \textbf{0.453} & \uline{0.487} & 0.993 & 1.604 & 1.603 & 1.803 \\ 
\hline 

\multirow{5}*{\rotatebox{90}{Weather}} & 24 & \uline{0.088} & \textbf{0.087} & 0.324 & 0.484 & 0.572 & 0.647 \\ 
\multicolumn{1}{c|}{} & 48 & \uline{0.114} & \textbf{0.113} & 0.366 & 0.608 & 0.647 & 0.720 \\ 
\multicolumn{1}{c|}{} & 168& \textbf{0.177} & \uline{0.178} & 0.543 & 1.081 & 1.117 & 1.351 \\ 
\multicolumn{1}{c|}{} & 336 & \textbf{0.241} & \uline{0.244} & 0.568 & 1.654 & 1.783 & 2.019 \\ 
\multicolumn{1}{c|}{} & 720 & \textbf{0.319} & \uline{0.321} & 0.601 & 1.401 & 1.850 & 2.109 \\ 
\hline 

\multicolumn{3}{c|}{ \makecell{\textbf{\citsm-Best} \\ \textbf{\% improvement}}} & \textbf{2.3\%}	& \textbf{54.2\%} & \textbf{71.7\%} & \textbf{73.3\%} & \textbf{75.8\%} \\ \hline
\end{tabular}}
\caption{Forecasting via Representation Learning (MSE)}
\label{tab:fm_1}
\end{table}

\begin{table}[t]
    \setlength{\tabcolsep}{2pt}
    \centering
    \scalebox{0.9}{
    \begin{tabular}{c|c|c|c|c|c|c|c}
        \toprule
        
        & FL  & \multicolumn{3}{c|}{\makecell{\citsm-Best (SS) \\ Model Size: Big}}& \makecell{\textbf{\citsm-}\\\textbf{Overall}\\\textbf{-Best (SS)}} & \makecell{PatchTST \\ (SS)}& \makecell{\citsm-\\Best} \\ \hline

        & & SAME & ALL & TL & & & Supervised\\ \hline
        \multirow{4}*{\rotatebox{90}{Electricity}} &96 & \uline{0.127} & 0.128 & 0.129 & \uline{0.127} & \textbf{0.126} & 0.129 \\ 
        \multicolumn{1}{c|}{} & 192  & \textbf{0.145} & \uline{0.146} & 0.147 & \textbf{0.145} & \textbf{0.145} & \uline{0.146} \\ 
        \multicolumn{1}{c|}{} & 336 & \textbf{0.156} & \textbf{0.156} & 0.159 & \textbf{0.156} & 0.164 & \uline{0.158} \\ 
        \multicolumn{1}{c|}{} & 720 & 0.187 & \textbf{0.185} & 0.189 & \textbf{0.185} & 0.193 & \uline{0.186} \\ 
        \hline 
        \multirow{4}*{\rotatebox{90}{Traffic}} &96 & \uline{0.350} & \textbf{0.348} & \uline{0.350} & \textbf{0.348} & 0.352 & 0.356 \\ 
        \multicolumn{1}{c|}{} & 192  & 0.372 & \textbf{0.370} & 0.372 & \textbf{0.370} & \uline{0.371} & 0.377 \\ 
        \multicolumn{1}{c|}{} & 336  & \textbf{0.379} & \uline{0.380} & \textbf{0.379} & \textbf{0.379} & 0.381 & 0.385 \\ 
        \multicolumn{1}{c|}{} & 720 & \textbf{0.420} & \uline{0.421} & 0.422 & \textbf{0.420} & 0.425 & 0.424 \\ 
        \hline 
        \multirow{4}*{\rotatebox{90}{Weather}} &96 & \uline{0.144} & \textbf{0.143} & \uline{0.144} & \textbf{0.143} & \uline{0.144} & 0.146 \\ 
        \multicolumn{1}{c|}{} & 192 & \textbf{0.189} & \textbf{0.189} & \uline{0.190} & \textbf{0.189} & \uline{0.190} & 0.191 \\ 
        \multicolumn{1}{c|}{} & 336 & \uline{0.240} & \textbf{0.239} & 0.241 & \textbf{0.239} & 0.244 & 0.243 \\ 
        \multicolumn{1}{c|}{} & 720 & \textbf{0.311} & \uline{0.316} & 0.319 & \textbf{0.311} & 0.321 & \uline{0.316} \\ 
        \hline 
        \multicolumn{6}{c|}{\makecell{\textbf{\citsm-Overall-Best (SS)} \\\textbf{\% Improvement (MSE)}}} & \textbf{1.45\%}  & \textbf{1.41\%}  \\
        \hline 
    \end{tabular}
    }
    \caption{Self-Supervision(SS) with varying pretrain strategies (MSE). PatchTST (SS) results are reported from ~\cite{patchtst}(SS finetuning mode)}
    \label{tab:fm-2-all}
\end{table}

\begin{figure}[t]
     \centering
     \includegraphics[width=1\columnwidth]{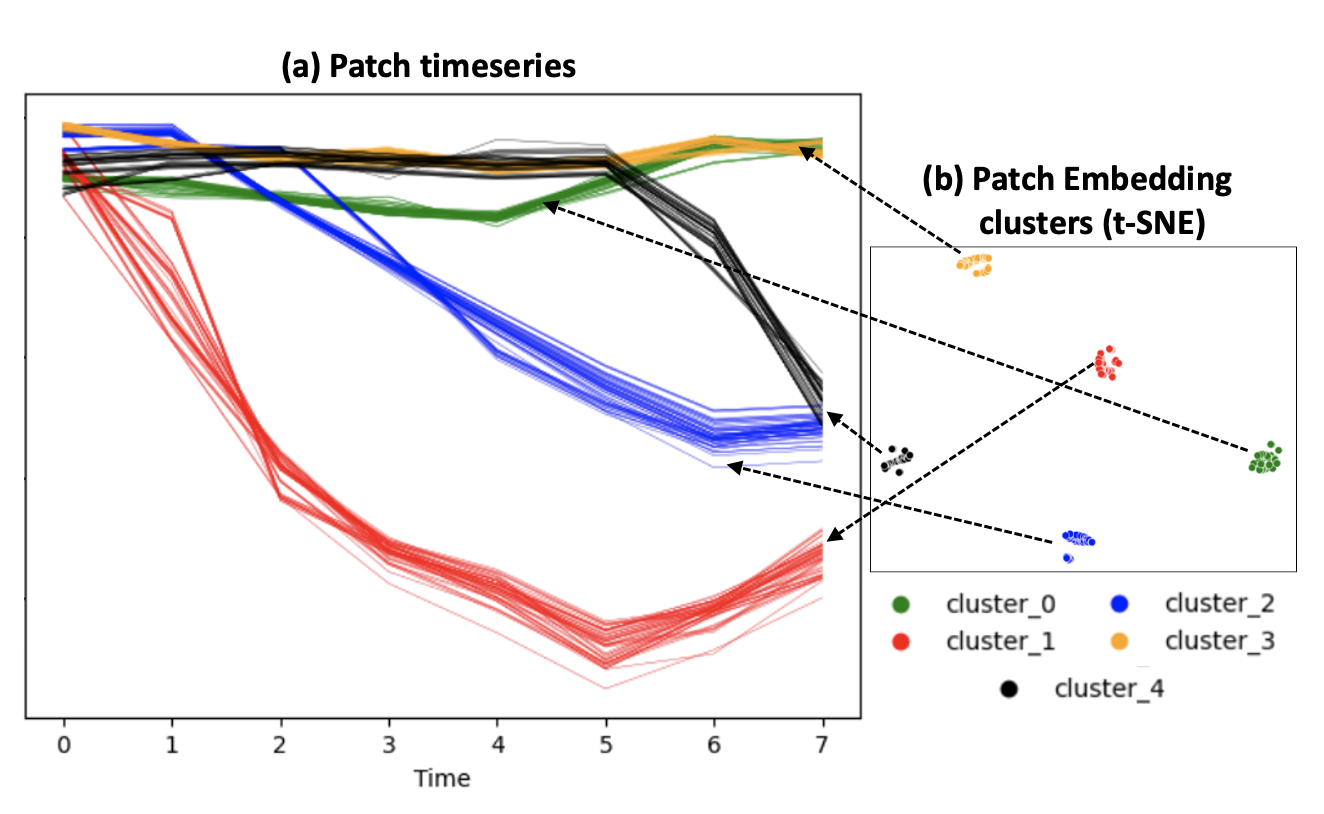}
    \caption{Correlation between Patch time-series and its associated embeddings.}
    \label{fig:patch_emb}
         
\end{figure}

\subsubsection{\textbf{Patch Representations}}
To understand the semantic meaning of the learned patch representations, we randomly choose 5 patch embeddings (i.e. output of ~\tsm~ backbone after pretraining) from ETTH1 and fetched its nearest 50 embeddings to form 5 clusters in the patch embedding space (Figure.~\ref{fig:patch_emb}(b)). We then fetch their associated patch time series and plot in Figure.~\ref{fig:patch_emb}(a). From the figure, we observe that nearby patch representations highly correlate to the patch time series of similar shapes and patterns, thereby learning meaningful patch representations that can effectively help in the finetuning process for various downstream tasks. 
Please refer to Appendix Figure.~\ref{fig:cluster_overall} for visualizations on more datasets.

\section{Conclusions and future directions}
Inspired by the success of MLP-Mixers in the vision domain, this paper proposes \textbf{~\tsm}, a purely designed MLP architecture with empirically validated time-series specific enhancements for multivariate forecasting and representation learning. Especially, we introduce a new hybrid architecture of augmenting various reconciliation heads and gated attention to the channel-independent backbone that significantly empowers the learning capability of simple MLP structures to outperform complex Transformer models. Through extensive experimentation, we show that ~\tsm~ outperforms all popular benchmarks with a significant reduction in compute resources. In future work, we plan to extend ~\tsm~ to other downstream tasks (such as classification, anomaly detection, etc.) and also improve the transfer learning capabilities across datasets. We also plan to investigate Swin\cite{hiremlp}, Shift\cite{shiftmlp}, and other newer Mixer variants\cite{cyclemlp}\cite{wavemlp} for its applicability in time-series.

\bibliographystyle{ACM-Reference-Format}
\bibliography{main}

\appendix

\section{Appendix}

\subsection{Datasets}

We use $7$ popular multivariate datasets provided in \citep{patchtst} for forecasting and representation learning. \textit{Weather}\footnote{https://www.bgc-jena.mpg.de/wetter/} dataset collects 21 meteorological indicators such as humidity and air temperature. \textit{Traffic}\footnote{https://pems.dot.ca.gov/} dataset reports the road occupancy rates from different sensors on San Francisco freeways. \textit{Electricity}\footnote{https://archive.ics.uci.edu/ml/datasets/ElectricityLoadDiagrams20112014} captures the hourly electricity consumption of 321 customers. \textit{ETT}\footnote{https://github.com/zhouhaoyi/ETDataset} (Electricity Transformer Temperature) datasets report sensor details from two electric Transformers in different resolutions (15 minutes and 1 hour). Thus, in total we have 4 ETT datasets: \textit{ETTM1}, \textit{ETTM2}, \textit{ETTH1}, and \textit{ETTH2}. 

\subsection{Supplementary Details}
TSMixer follows ~\cite{lr_sched} to determine the optimal learning rate during training for better convergence. Early stopping with patience 10 is applied during training. TSMixer library is built using PyTorch and multi-GPU node training is enabled via Pytorch DDP\footnote{$https://pytorch.org/tutorials/beginner/dist\_overview.html$}. TSMixer can easily scale and cater to the needs of large-scale forecasting and can be deployed across multiple nodes in the Kubernetes clusters.

\subsection{Supplementary Figures}
\begin{figure}[!h]
     \centering
     \begin{subfigure}[b]{0.49\columnwidth}
         \centering
         \includegraphics[width=0.8\columnwidth]{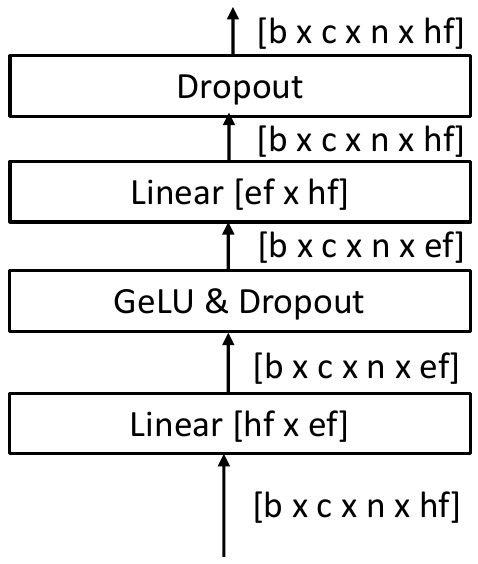}
         \caption{MLP block in \mixer}
         \label{fig:mlp}
     \end{subfigure}
     \begin{subfigure}[b]{0.49\columnwidth}
         \centering
         \includegraphics[width=0.8\columnwidth]{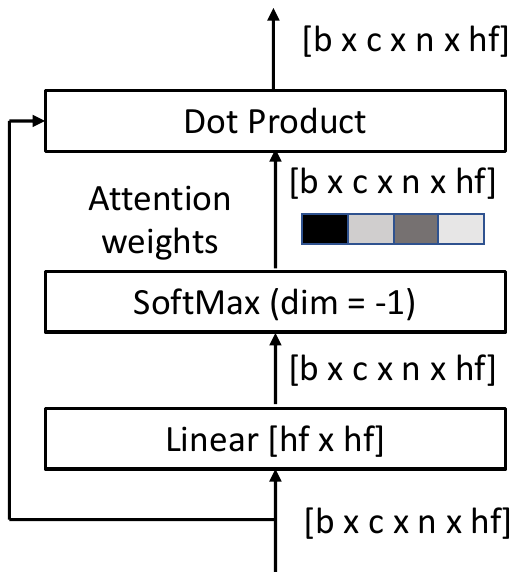}
         \caption{Gated attention in ours}
         \label{fig:ga}
     \end{subfigure}
    \caption{
    MLP and Gated Attention 
    }
    \label{fig:gated-attention}
\end{figure}

\begin{figure}[!h]
     \centering
     \begin{subfigure}[b]{0.49\columnwidth}
         \centering
         \includegraphics[width=0.8\columnwidth]{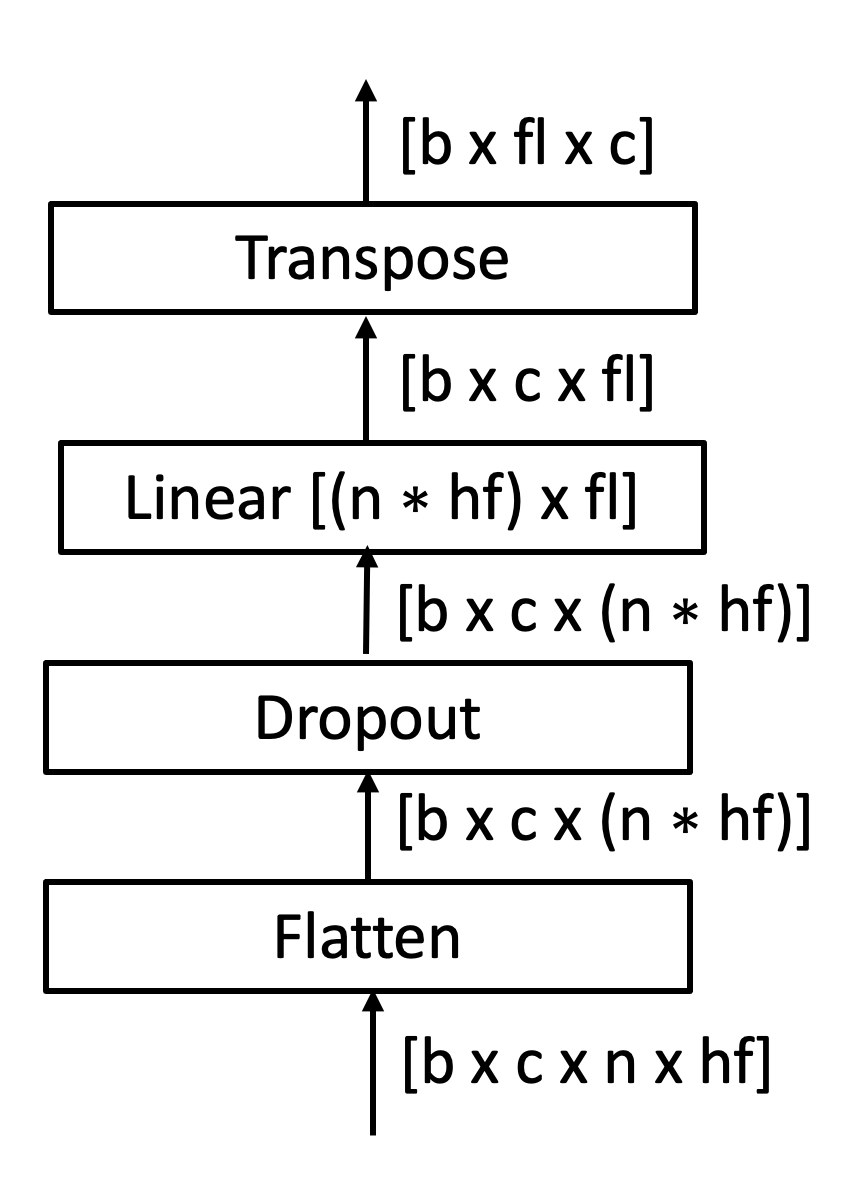}
         \caption{Prediction head}
         \label{fig:model-heads:prediction}
     \end{subfigure}
     \begin{subfigure}[b]{0.49\columnwidth}
         \centering
         \includegraphics[width=0.8\columnwidth]{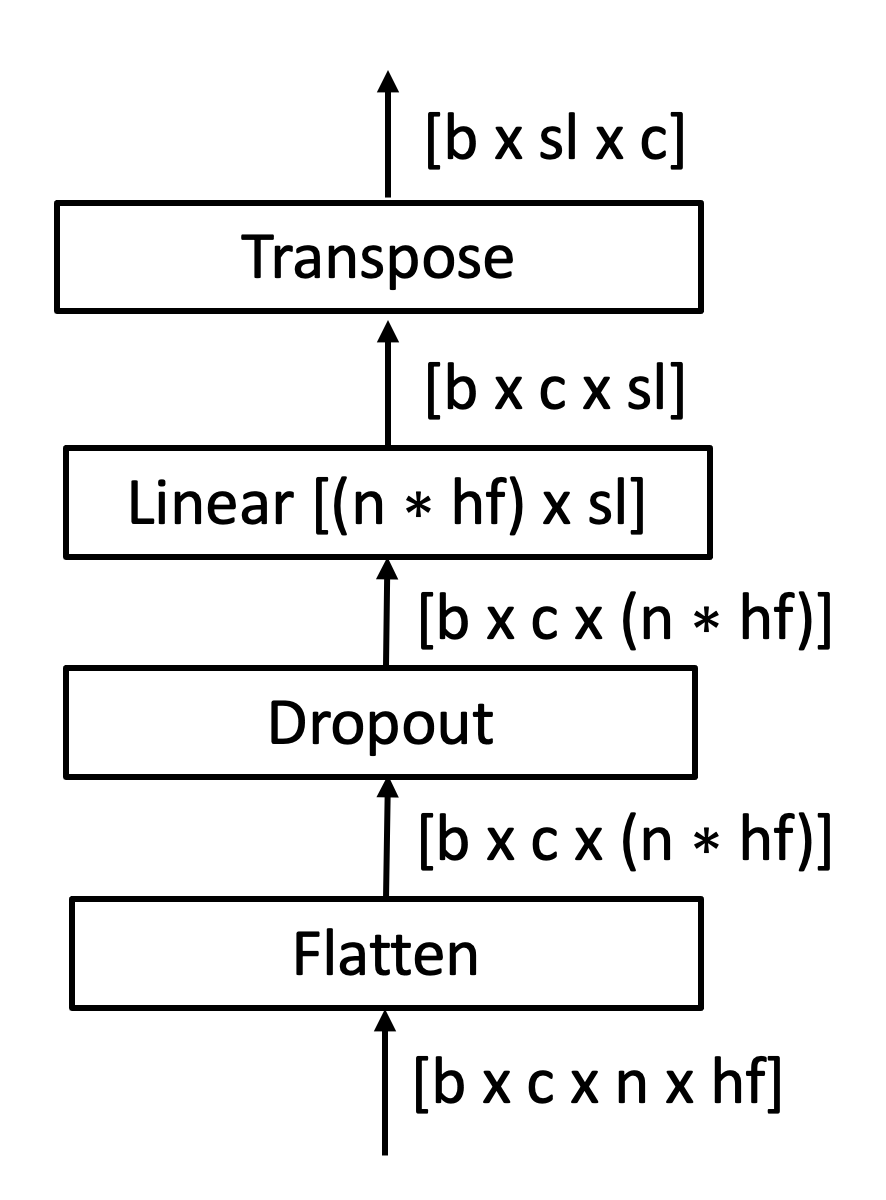}
         \caption{Pretrain head}
         \label{fig:model-heads:pretrain}
     \end{subfigure}
    \caption{
    Model heads for the two workflows.
    }
    \label{fig:model-heads}
\end{figure}

This section covers the supplementary figures that are referred in the paper for better understanding. Figure.~\ref{fig:gated-attention}(a) depicts the standard MLP block used in the mixer layer. Figure.~\ref{fig:gated-attention}(b) explains the gated attention block used in the mixer layer to downscale the noisy features. Figure.~\ref{fig:model-heads} highlights the architecture followed in the prediction and pretrain head. In the self-supervised pre-training workflow, pre-train head is attached to the backbone. In other workflows (such as supervised, or finetuning), the prediction head is attached. 

\begin{figure*}[!t]
    \centering
    \captionsetup{width=.8\linewidth}
    \includegraphics[scale=0.5]{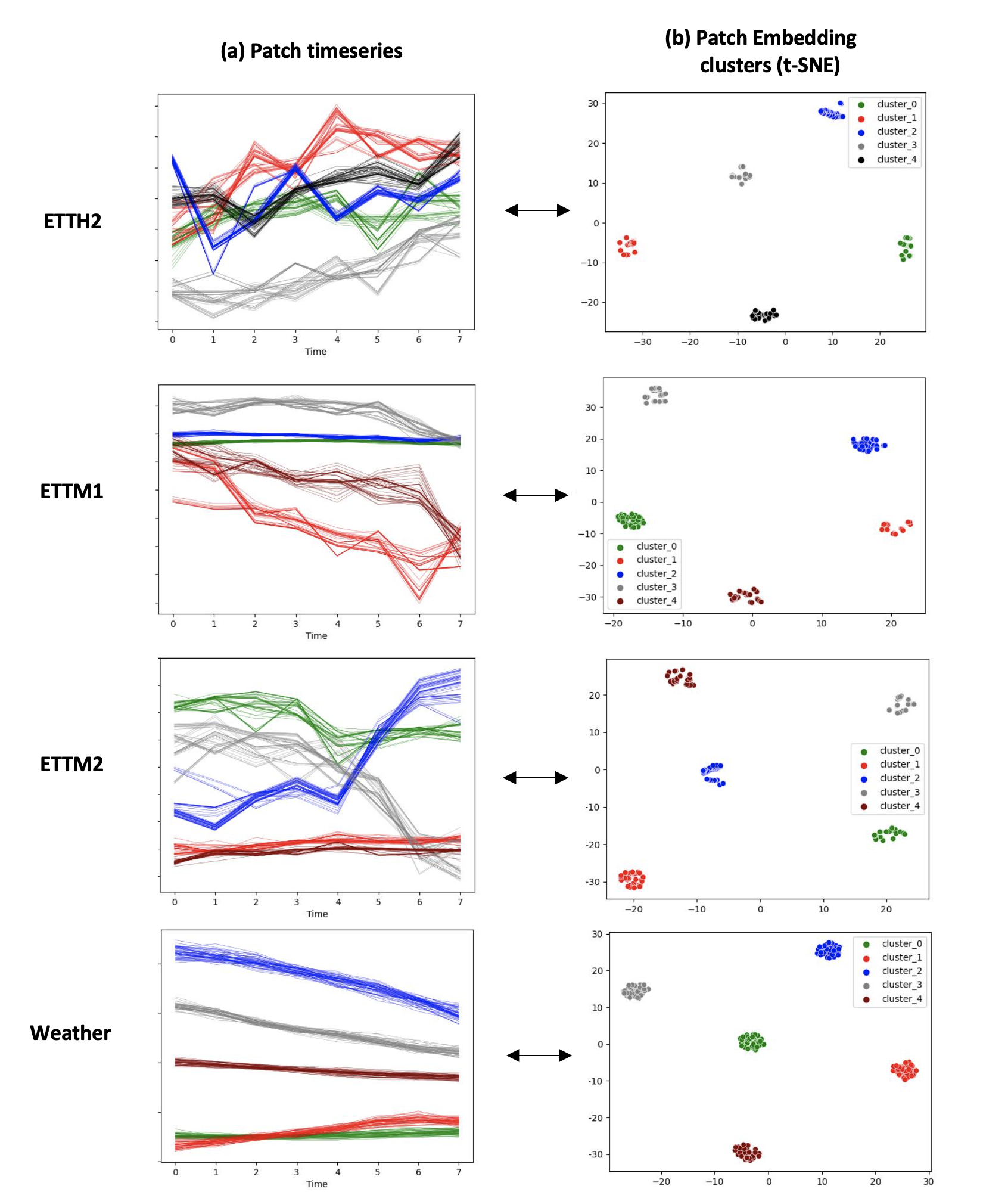}
    \caption{Correlation between Patch time-series and its associated embeddings in multiple datasets. Nearby patch representations highly correlate to the patch time series of similar shapes and patterns, thereby learning meaningful patch representations.}
    \label{fig:cluster_overall}
\end{figure*}

Figure.~\ref{fig:cluster_overall} shows the correlation between patch time series and its associated embeddings in different datasets.

\subsection{Supplementary results}

This section explains the supplementary results which are not reported in the main paper due to space constraints.

\subsubsection{\textbf{Benchmarking LightTS, S4, CrossFormer and TS2Vec}}
\label{appendix_1}
This section compares and contrasts TSMixer with LightTS~\cite{lightts}, S4~\cite{s4}, CrossFormer~\cite{crossformer} and TS2Vec~\cite{ts2vec}. Considering the space constraints and the lower performance of these benchmarks as compared to our reported primary benchmarks (like PatchTST, DLinear), we mention these in the appendix instead of the main paper.
Table.~\ref{tab:lightts} and Table.~\ref{tab:crossformer} compare and contrast TSMixer with LightTS, S4 and CrossFormer in a supervised workflow. Table.~\ref{tab:ab_ts2vec} compares and contrasts TSMixer with TS2Vec in the self-supervised workflow. Since the baseline papers reported the results in a different forecast horizon ($fl$) space, we report their comparison in separate tables as per the commonly available forecast horizons.

\subsubsection{\textbf{Detailed Analysis}}
In this section, we provide more detailed benchmarking results as follows:
\begin{itemize}
    \item Table.~\ref{tab:supervised-ab1} shows the drill-down view of ~\tsmb~ on supervised multivariate forecasting.
    \item Table.~\ref{tab:ga_hc-ab1} highlights the effect of CI, Gated Attention and Hierarchy Reconciliation over Vanilla TSMixer on all datasets.
    \item Table.~\ref{tab:cc-detail} depicts the MSE analysis of various channel mixing techniques with different context lengths.
    \item Table.~\ref{tab:ab_fm_1} highlights the MSE drill-down view of ~\tsmb~ for self-supervised multivariate forecasting.
\end{itemize}

\begin{table}
\centering
\setlength{\tabcolsep}{2pt}
\scalebox{0.9}{
\begin{tabular}{c|c|c|c|c}
\hline  
Data & \makecell{FL} & \makecell{\textbf{\citsm-Best}}  & LightTS & S4  \\
\hline
\multirow{2}{*}{ETTH1} & 336 & \textbf{0.421} & 0.466 & 1.407  \\ 
& 720 & \textbf{0.444}  & 0.542 & 1.162 \\ \hline
\multirow{2}{*}{ETTH2} & 336 & \textbf{0.357} & 0.497 & 1.98  \\ 
& 720 & \textbf{0.395} & 0.739 & 2.65  \\ \hline
\multirow{2}{*}{Weather} & 336 & \textbf{0.243} & 0.527 & 0.531  \\ 
& 720 & \textbf{0.316} & 0.554 & 0.578  \\ \hline
\multicolumn{3}{c|}{\makecell{\textbf{\citsm-Best \% improvement }}}& \textbf{33.2\%} & \textbf{66.4\%} \\ \hline
\end{tabular}}
\caption{LightTS and S4 MSE Benchmarking - Baseline results reported from ~\cite{lightts} and ~\cite{s4}} 
\label{tab:lightts}
\end{table}

\begin{table}
\centering
\setlength{\tabcolsep}{2pt}
\scalebox{0.9}{
\begin{tabular}{c|c|c|c|c}
\hline  
Data & \makecell{FL} & CrossFormer & \makecell{\citsm} & \makecell{\textbf{\citsm-Best}}  \\
\hline
\multirow{2}*{{ETTH1}} & 336 & 0.44 & \uline{0.437} & \textbf{0.421} \\ 
\multicolumn{1}{c|}{} & 720 & 0.519 & \uline{0.465} & \textbf{0.444} \\ \hline
\multirow{2}*{{Electricity}} & 336 & 0.323 & \uline{0.165} & \textbf{0.158} \\ 
\multicolumn{1}{c|}{} & 720 & 0.404 & \uline{0.204} & \textbf{0.186} \\ 
\hline 
\multirow{2}*{{Traffic}} & 336 & 0.53 & \uline{0.388} & \textbf{0.385} \\ 
\multicolumn{1}{c|}{} & 720 & 0.573 & \uline{0.426} & \textbf{0.424} \\ \hline
\multirow{2}*{{Weather}} & 336 & 0.495 & \uline{0.246} & \textbf{0.243} \\ 
\multicolumn{1}{c|}{} & 720 & 0.526 & \uline{0.323} & \textbf{0.316} \\ 
\hline 
\multicolumn{3}{c|}{\makecell{\textbf{\% Improvement} \\ \textbf{over CrossFormer}}}& \textbf{31.35\%} & \textbf{33.50\%} \\ \hline
\end{tabular}}
\caption{CrossFormer MSE Benchmarking - Baseline results reported from ~\cite{crossformer}}
\label{tab:crossformer}
\end{table}

\begin{table*}
\centering
\setlength{\tabcolsep}{2pt}
\scalebox{0.9}{
\begin{tabular}{c|c|c|c}
\hline  
& \makecell{FL} & \makecell{CI-\tsm\\-Best} & \makecell{TS2Vec} \\
\hline
\multirow{5}*{ETTH1} & 24 & \textbf{0.314} & {0.599} \\ 
\multicolumn{1}{c|}{} & 48 & \textbf{0.343} & {0.629} \\ 
\multicolumn{1}{c|}{} & 168& \textbf{0.397} & {0.755} \\ 
\multicolumn{1}{c|}{} & 336 & \textbf{0.424} & {0.907} \\ 
\multicolumn{1}{c|}{} & 720 & \textbf{0.453} & {1.048} \\ 
\hline 
\multirow{2}*{Electricity} & 336 & \textbf{0.16} & {0.349} \\ 
\multicolumn{1}{c|}{} & 720 & \textbf{0.187} & {0.375} \\ \hline
&\multicolumn{2}{c|}{\makecell{\textbf{\% Improvement}}}	& \textbf{50.7\%} \\ \hline

\end{tabular}}
\caption{TS2Vec MSE Benchmarking - Baseline results reported from ~\cite{ts2vec}}
\label{tab:ab_ts2vec}
\end{table*}

\begin{table*}[h!]
	\centering
        \captionsetup{width=.8\linewidth}
        \scalebox{0.9}{
		\begin{tabular}{cc|c|cc|cc|cc|cc|ccc}
			\cline{2-13}
			&\multicolumn{2}{c|}{Models}& \multicolumn{2}{c|}{PatchTST}& \multicolumn{2}{c|}{DLinear}& \multicolumn{2}{c|}{\citsm(\ga,\hr)}& \multicolumn{2}{c|}{\citsm(\ga,\hr,\cc)}& \multicolumn{2}{c}{\citsm-Best} \\
			\cline{2-13}
			&\multicolumn{2}{c|}{Metric}&MSE&MAE&MSE&MAE&MSE&MAE&MSE&MAE&MSE&MAE\\
			\cline{2-13}
        
            &\multirow{4}*{\rotatebox{90}{ETTH1}} & 96 & \uline{0.37} & 0.4 & 0.375 & \uline{0.399} & \textbf{0.368} & \textbf{0.398} & \textbf{0.368} & \textbf{0.398} & \textbf{0.368} & \textbf{0.398} \\ 
            &\multicolumn{1}{c|}{} & 192 & 0.413 & 0.429 & 0.405 & \textbf{0.416} & \textbf{0.399} & \uline{0.418} & \uline{0.4} & \uline{0.418} & \textbf{0.399} & \uline{0.418} \\ 
            &\multicolumn{1}{c|}{} & 336 & \uline{0.422} & 0.44 & 0.439 & 0.443 & \textbf{0.421} & \textbf{0.436} & \uline{0.422} & \uline{0.437} & \textbf{0.421} & \textbf{0.436} \\ 
            &\multicolumn{1}{c|}{} & 720 & \uline{0.447} & \uline{0.468} & 0.472 & 0.49 & \textbf{0.444} & \textbf{0.467} & 0.45 & \textbf{0.467} & \textbf{0.444} & \textbf{0.467} \\ 
            \cline{2-13} 
            &\multirow{4}*{\rotatebox{90}{ETTH2}} & 96 & \textbf{0.274} & \textbf{0.337} & 0.289 & 0.353 & \uline{0.276} & \textbf{0.337} & \uline{0.276} & \uline{0.339} & \uline{0.276} & \textbf{0.337} \\ 
            &\multicolumn{1}{c|}{} & 192 & 0.341 & 0.382 & 0.383 & 0.418 & \uline{0.335} & \uline{0.377} & \textbf{0.33} & \textbf{0.374} & \textbf{0.33} & \textbf{0.374} \\ 
            &\multicolumn{1}{c|}{} & 336 & \textbf{0.329} & \textbf{0.384} & 0.448 & 0.465 & 0.369 & 0.406 & \uline{0.357} & \uline{0.401} & \uline{0.357} & \uline{0.401} \\ 
            &\multicolumn{1}{c|}{} & 720 & \textbf{0.379} & \textbf{0.422} & 0.605 & 0.551 & 0.409 & 0.447 & \uline{0.395} & \uline{0.436} & \uline{0.395} & \uline{0.436} \\ 
            \cline{2-13} 
            &\multirow{4}*{\rotatebox{90}{ETTM1}} & 96 & \uline{0.293} & \uline{0.346} & 0.299 & \textbf{0.343} & 0.297 & 0.348 & \textbf{0.291} & \uline{0.346} & \textbf{0.291} & \uline{0.346} \\ 
            &\multicolumn{1}{c|}{} & 192 & \textbf{0.333} & 0.37 & 0.335 & \textbf{0.365} & \textbf{0.333} & \uline{0.369} & \uline{0.334} & \uline{0.369} & \textbf{0.333} & \uline{0.369} \\ 
            &\multicolumn{1}{c|}{} & 336 & 0.369 & 0.392 & 0.369 & \uline{0.386} & \textbf{0.365} & \textbf{0.385} & \uline{0.367} & 0.387 & \textbf{0.365} & \textbf{0.385} \\ 
            &\multicolumn{1}{c|}{} & 720 & \textbf{0.416} & 0.42 & 0.425 & 0.421 & \textbf{0.416} & \textbf{0.413} & \uline{0.421} & \uline{0.415} & \textbf{0.416} & \textbf{0.413} \\ 
            \cline{2-13} 
            &\multirow{4}*{\rotatebox{90}{ETTM2}} & 96 & 0.166 & \uline{0.256} & 0.167 & 0.26 & \textbf{0.164} & \textbf{0.255} & \uline{0.165} & \textbf{0.255} & \textbf{0.164} & \textbf{0.255} \\ 
            &\multicolumn{1}{c|}{} & 192 & \uline{0.223} & \uline{0.296} & 0.224 & 0.303 & \textbf{0.219} & \textbf{0.293} & 0.225 & 0.299 & \textbf{0.219} & \textbf{0.293} \\ 
            &\multicolumn{1}{c|}{} & 336 & \uline{0.274} & \textbf{0.329} & 0.281 & 0.342 & \textbf{0.273} & \uline{0.33} & \textbf{0.273} & \textbf{0.329} & \textbf{0.273} & \textbf{0.329} \\ 
            &\multicolumn{1}{c|}{} & 720 & 0.362 & 0.385 & 0.397 & 0.421 & \textbf{0.358} & \textbf{0.38} & \uline{0.361} & \uline{0.383} & \textbf{0.358} & \textbf{0.38} \\ 
            \cline{2-13} 
            &\multirow{4}*{\rotatebox{90}{Electricity}} & 96 & \textbf{0.129} & \textbf{0.222} & \uline{0.14} & 0.237 & \textbf{0.129} & \uline{0.224} & \textbf{0.129} & 0.225 & \textbf{0.129} & \uline{0.224} \\ 
            &\multicolumn{1}{c|}{} & 192 & \uline{0.147} & \textbf{0.24} & 0.153 & 0.249 & 0.148 & \uline{0.242} & \textbf{0.146} & \uline{0.242} & \textbf{0.146} & \uline{0.242} \\ 
            &\multicolumn{1}{c|}{} & 336 & \uline{0.163} & \uline{0.259} & 0.169 & 0.267 & 0.164 & \uline{0.259} & \textbf{0.158} & \textbf{0.256} & \textbf{0.158} & \textbf{0.256} \\ 
            &\multicolumn{1}{c|}{} & 720 & \uline{0.197} & \uline{0.29} & 0.203 & 0.301 & 0.201 & 0.292 & \textbf{0.186} & \textbf{0.282} & \textbf{0.186} & \textbf{0.282} \\ 
            \cline{2-13} 
            &\multirow{4}*{\rotatebox{90}{Traffic}} & 96 & \uline{0.36} & \uline{0.249} & 0.41 & 0.282 & \textbf{0.356} & \textbf{0.248} & 0.369 & 0.264 & \textbf{0.356} & \textbf{0.248} \\ 
            &\multicolumn{1}{c|}{} & 192 & \uline{0.379} & \textbf{0.256} & 0.423 & 0.287 & \textbf{0.377} & \uline{0.257} & 0.393 & 0.278 & \textbf{0.377} & \uline{0.257} \\ 
            &\multicolumn{1}{c|}{} & 336 & \uline{0.392} & \uline{0.264} & 0.436 & 0.296 & \textbf{0.385} & \textbf{0.262} & 0.406 & 0.285 & \textbf{0.385} & \textbf{0.262} \\ 
            &\multicolumn{1}{c|}{} & 720 & \uline{0.432} & \uline{0.286} & 0.466 & 0.315 & \textbf{0.424} & \textbf{0.283} & 0.445 & 0.304 & \textbf{0.424} & \textbf{0.283} \\ 
            \cline{2-13} 
            &\multirow{4}*{\rotatebox{90}{Weather}} & 96 & 0.149 & \uline{0.198} & 0.176 & 0.237 & \uline{0.148} & \uline{0.198} & \textbf{0.146} & \textbf{0.197} & \textbf{0.146} & \textbf{0.197} \\ 
            &\multicolumn{1}{c|}{} & 192 & 0.194 & \uline{0.241} & 0.22 & 0.282 & \uline{0.193} & \textbf{0.24} & \textbf{0.191} & \textbf{0.24} & \textbf{0.191} & \textbf{0.24} \\ 
            &\multicolumn{1}{c|}{} & 336 & 0.245 & 0.282 & 0.265 & 0.319 & \textbf{0.243} & \textbf{0.279} & \uline{0.244} & \uline{0.28} & \textbf{0.243} & \textbf{0.279} \\ 
            &\multicolumn{1}{c|}{} & 720 & \textbf{0.314} & \uline{0.334} & 0.323 & 0.362 & 0.317 & \textbf{0.333} & \uline{0.316} & \textbf{0.333} & \uline{0.316} & \textbf{0.333} \\ 
            \cline{2-13} 
		\end{tabular}
	}
	\caption{Drill-down view of ~\tsmb~ for supervised multivariate forecasting.~\tsmb~is defined as the best of ~\tsmgh~and~\tsmghc, wherein. based on the considered dataset - either ~\tsmgh~ or ~\tsmghc~ outperforms the existing benchmarks. ~\tsmghc~ uses context length = 1 for this experiment. PatchTST results are reported from ~\cite{patchtst} and DLinear from ~\cite{dlinear}}
	\label{tab:supervised-ab1}
\end{table*}

\begin{table*}[t]
	\centering
        \captionsetup{width=.8\linewidth}
        \scalebox{0.9}{
		\begin{tabular}{cc|c|cc|cc|cc|cc|ccc}
			\cline{2-13}
			&\multicolumn{2}{c|}{Models}& \multicolumn{2}{c|}{V-\tsm}& \multicolumn{2}{c|}{CI-\tsm}& \multicolumn{2}{c|}{\citsm(\ga)}& \multicolumn{2}{c|}{\citsm(\hr)}& \multicolumn{2}{c}{\citsm(\ga,\hr)} \\
			\cline{2-13}
			&\multicolumn{2}{c|}{Metric}&MSE&MAE&MSE&MAE&MSE&MAE&MSE&MAE&MSE&MAE\\
			\cline{2-13}
        &\multirow{4}*{\rotatebox{90}{ETTH1}} &96& 0.449 & 0.462 & \uline{0.375} & 0.401 & \uline{0.375} & \uline{0.4} & 0.377 & 0.405 & \textbf{0.368} & \textbf{0.398} \\ 
        &\multicolumn{1}{c|}{} & 192 & 0.485 & 0.484 & 0.411 & 0.426 & \uline{0.408} & \uline{0.422} & 0.41 & 0.428 & \textbf{0.399} & \textbf{0.418} \\ 
        &\multicolumn{1}{c|}{} & 336 & 0.504 & 0.497 & 0.437 & 0.445 & 0.433 & \uline{0.439} & \uline{0.431} & 0.445 & \textbf{0.421} & \textbf{0.436} \\ 
        &\multicolumn{1}{c|}{} & 720 & 0.573 & 0.534 & 0.465 & 0.478 & \uline{0.454} & \uline{0.47} & 0.457 & 0.477 & \textbf{0.444} & \textbf{0.467} \\ 
        \cline{2-13} 
        &\multirow{4}*{\rotatebox{90}{ETTH2}} &96& 0.369 & 0.412 & 0.284 & 0.342 & \textbf{0.276} & \uline{0.339} & \uline{0.283} & 0.341 & \textbf{0.276} & \textbf{0.337} \\ 
        &\multicolumn{1}{c|}{} & 192 & 0.391 & 0.428 & 0.353 & 0.384 & \uline{0.338} & \uline{0.379} & 0.344 & 0.38 & \textbf{0.335} & \textbf{0.377} \\ 
        &\multicolumn{1}{c|}{} & 336 & 0.403 & 0.44 & \uline{0.365} & \textbf{0.403} & \textbf{0.362} & \uline{0.404} & 0.376 & 0.408 & 0.369 & 0.406 \\ 
        &\multicolumn{1}{c|}{} & 720 & 0.475 & 0.481 & \uline{0.406} & \uline{0.442} & 0.415 & 0.451 & \textbf{0.396} & \textbf{0.434} & 0.409 & 0.447 \\ 
        \cline{2-13} 
        &\multirow{4}*{\rotatebox{90}{ETTM1}} &96& 0.328 & 0.372 & 0.305 & 0.354 & \textbf{0.296} & \uline{0.349} & 0.301 & 0.351 & \uline{0.297} & \textbf{0.348} \\ 
        &\multicolumn{1}{c|}{} & 192 & 0.372 & 0.397 & 0.336 & 0.374 & \uline{0.334} & \uline{0.37} & 0.338 & 0.372 & \textbf{0.333} & \textbf{0.369} \\ 
        &\multicolumn{1}{c|}{} & 336 & 0.405 & 0.416 & 0.375 & 0.398 & \uline{0.366} & \uline{0.387} & 0.376 & 0.394 & \textbf{0.365} & \textbf{0.385} \\ 
        &\multicolumn{1}{c|}{} & 720 & 0.457 & 0.445 & 0.425 & 0.417 & \uline{0.419} & \uline{0.415} & 0.422 & 0.417 & \textbf{0.416} & \textbf{0.413} \\ 
        \cline{2-13} 
        &\multirow{4}*{\rotatebox{90}{ETTM2}} &96& 0.195 & 0.282 & 0.172 & 0.263 & 0.175 & 0.264 & \uline{0.168} & \uline{0.259} & \textbf{0.164} & \textbf{0.255} \\ 
        &\multicolumn{1}{c|}{} & 192 & 0.26 & 0.326 & \uline{0.221} & \uline{0.294} & \uline{0.221} & \uline{0.294} & 0.224 & 0.296 & \textbf{0.219} & \textbf{0.293} \\ 
        &\multicolumn{1}{c|}{} & 336 & 0.341 & 0.373 & \textbf{0.273} & \textbf{0.328} & 0.277 & 0.333 & \uline{0.276} & 0.331 & \textbf{0.273} & \uline{0.33} \\ 
        &\multicolumn{1}{c|}{} & 720 & 0.437 & 0.432 & \textbf{0.357} & 0.384 & 0.365 & 0.385 & 0.36 & \uline{0.383} & \uline{0.358} & \textbf{0.38} \\ 
        \cline{2-13} 
        &\multirow{4}*{\rotatebox{90}{Electricity}} &96& 0.212 & 0.323 & \uline{0.13} & \uline{0.224} & \textbf{0.129} & \textbf{0.223} & \uline{0.13} & 0.226 & \textbf{0.129} & \uline{0.224} \\ 
        &\multicolumn{1}{c|}{} & 192 & \uline{0.21} & 0.319 & \textbf{0.148} & \uline{0.242} & \textbf{0.148} & \textbf{0.241} & \textbf{0.148} & 0.243 & \textbf{0.148} & \uline{0.242} \\ 
        &\multicolumn{1}{c|}{} & 336 & 0.224 & 0.332 & \uline{0.165} & \textbf{0.259} & \uline{0.165} & \textbf{0.259} & \uline{0.165} & \uline{0.26} & \textbf{0.164} & \textbf{0.259} \\ 
        &\multicolumn{1}{c|}{} & 720 & 0.251 & 0.352 & \uline{0.204} & 0.293 & \textbf{0.201} & \textbf{0.291} & \uline{0.204} & 0.295 & \textbf{0.201} & \uline{0.292} \\ 
        \cline{2-13} 
        &\multirow{4}*{\rotatebox{90}{Traffic}} &96& 0.618 & 0.386 & 0.358 & 0.251 & \textbf{0.355} & \textbf{0.246} & 0.357 & 0.251 & \uline{0.356} & \uline{0.248} \\ 
        &\multicolumn{1}{c|}{} & 192 & 0.624 & 0.385 & 0.379 & \uline{0.26} & \textbf{0.377} & \textbf{0.257} & \uline{0.378} & \uline{0.26} & \textbf{0.377} & \textbf{0.257} \\ 
        &\multicolumn{1}{c|}{} & 336 & 0.63 & 0.38 & 0.388 & 0.265 & \uline{0.387} & \uline{0.264} & \uline{0.387} & 0.265 & \textbf{0.385} & \textbf{0.262} \\ 
        &\multicolumn{1}{c|}{} & 720 & 0.66 & 0.389 & 0.426 & 0.286 & 0.427 & \uline{0.285} & \uline{0.425} & 0.286 & \textbf{0.424} & \textbf{0.283} \\ 
        \cline{2-13} 
        &\multirow{4}*{\rotatebox{90}{Weather}} &96& 0.159 & 0.216 & 0.15 & 0.202 & \uline{0.149} & \uline{0.199} & 0.151 & 0.202 & \textbf{0.148} & \textbf{0.198} \\ 
        &\multicolumn{1}{c|}{} & 192 & 0.207 & 0.257 & \uline{0.195} & 0.244 & \uline{0.195} & \uline{0.242} & \uline{0.195} & 0.243 & \textbf{0.193} & \textbf{0.24} \\ 
        &\multicolumn{1}{c|}{} & 336 & 0.256 & 0.294 & \uline{0.246} & 0.284 & \uline{0.246} & \uline{0.281} & \uline{0.246} & 0.283 & \textbf{0.243} & \textbf{0.279} \\ 
        &\multicolumn{1}{c|}{} & 720 & 0.33 & 0.344 & 0.323 & 0.338 & \textbf{0.317} & \textbf{0.333} & \uline{0.321} & \uline{0.337} & \textbf{0.317} & \textbf{0.333} \\ 
        \cline{2-13} 
        \multicolumn{5}{c|}{ \makecell{\textbf{\% MSE improvement}  \textbf{over \vtsm}}} &\multicolumn{2}{c|}{\textbf{18\%}} &\multicolumn{2}{c|}{\textbf{18.5\%}}  &\multicolumn{2}{c|}{\textbf{18.1\%}} &\multicolumn{2}{c}{\textbf{19.3\%}} \\ 
            \cline{2-13} 
		\end{tabular}
	}
	\caption{Effect of CI, Gated Attention and Hierarchy Reconciliation over Vanilla TSMixer on all datasets. ~\citsm~outperforms ~\vtsm~ by 18\% and by adding gated attention (G) and Hierarchy Reconciliation head (H), the accuracy further improves by 1.3\%, leading to a total of 19.3\% improvement. It is important to note that, we observe stable improvements when  (G) and (H) are used together instead of just (G) or (H).}
	\label{tab:ga_hc-ab1}
\end{table*}

\begin{table*}[t]
    \captionsetup{width=.8\linewidth}
    \centering
    \scalebox{0.9}{
    \begin{tabular}{c|c|c|c|c|c|c|c|c|c|c}
        \toprule
        
        \multicolumn{2}{c|}{Models}& V-\tsm & CI-\tsm & IC-\tsm & \multicolumn{6}{c}{\citsm(\ga,\cc)} \\ \hline
        \multicolumn{2}{c|}{Channel Context Length}& \multicolumn{3}{c|}{} & (1) & (2) & (3) & (4) & (5) & (Best) \\ \hline
        \multirow{4}*{\rotatebox{90}{ETTH1}} &96& 0.449 & 0.375 & 0.379 & \textbf{0.373} & \textbf{0.373} & 0.377 & \textbf{0.373} & \uline{0.374} & \textbf{0.373} \\ 
        \multicolumn{1}{c|}{} & 192 & 0.485 & 0.411 & 0.416 & 0.416 & \textbf{0.407} & \textbf{0.407} & 0.409 & \uline{0.408} & \textbf{0.407} \\ 
        \multicolumn{1}{c|}{} & 336 & 0.504 & 0.437 & 0.437 & 0.436 & \uline{0.435} & 0.437 & 0.436 & \textbf{0.43} & \textbf{0.43} \\ 
        \multicolumn{1}{c|}{} & 720 & 0.573 & \uline{0.465} & 0.471 & 0.482 & 0.493 & 0.472 & \uline{0.465} & \textbf{0.454} & \textbf{0.454} \\ 
        \hline 
        \multirow{4}*{\rotatebox{90}{ETTH2}} &96& 0.369 & 0.284 & 0.291 & \uline{0.27} & 0.274 & 0.277 & 0.275 & \textbf{0.269} & \textbf{0.269} \\ 
        \multicolumn{1}{c|}{} & 192 & 0.391 & 0.353 & 0.345 & 0.336 & 0.347 & \uline{0.333} & \textbf{0.33} & 0.339 & \textbf{0.33} \\ 
        \multicolumn{1}{c|}{} & 336 & 0.403 & 0.365 & 0.361 & 0.364 & 0.363 & \uline{0.36} & \textbf{0.359} & 0.363 & \textbf{0.359} \\ 
        \multicolumn{1}{c|}{} & 720 & 0.475 & 0.406 & 0.419 & 0.4 & \uline{0.398} & 0.41 & \textbf{0.393} & \textbf{0.393} & \textbf{0.393} \\ 
        \hline 
        \multirow{4}*{\rotatebox{90}{ETTM1}} &96& 0.328 & 0.305 & 0.307 & \textbf{0.3} & \uline{0.302} & 0.304 & 0.309 & 0.305 & \textbf{0.3} \\ 
        \multicolumn{1}{c|}{} & 192 & 0.372 & \textbf{0.336} & 0.34 & 0.345 & 0.345 & 0.341 & 0.345 & \uline{0.338} & \uline{0.338} \\ 
        \multicolumn{1}{c|}{} & 336 & 0.405 & \uline{0.375} & \textbf{0.372} & 0.379 & \uline{0.375} & \uline{0.375} & 0.377 & 0.383 & \uline{0.375} \\ 
        \multicolumn{1}{c|}{} & 720 & 0.457 & \uline{0.425} & 0.428 & \textbf{0.422} & 0.426 & 0.429 & 0.426 & 0.44 & \textbf{0.422} \\ 
        \hline 
        \multirow{4}*{\rotatebox{90}{ETTM2}} &96& 0.195 & 0.172 & 0.169 & \uline{0.166} & 0.173 & \textbf{0.165} & \uline{0.166} & 0.168 & \textbf{0.165} \\ 
        \multicolumn{1}{c|}{} & 192 & 0.26 & 0.221 & 0.221 & 0.225 & \textbf{0.218} & \uline{0.22} & \uline{0.22} & 0.221 & \textbf{0.218} \\ 
        \multicolumn{1}{c|}{} & 336 & 0.341 & \uline{0.273} & 0.278 & \textbf{0.272} & 0.284 & 0.277 & 0.285 & 0.276 & \textbf{0.272} \\ 
        \multicolumn{1}{c|}{} & 720 & 0.437 & \textbf{0.357} & 0.367 & 0.359 & \uline{0.358} & 0.385 & 0.362 & 0.362 & \uline{0.358} \\ 
        \hline 
        \multirow{4}*{\rotatebox{90}{Electricity}} &96& 0.212 & \textbf{0.13} & 0.163 & 0.139 & \uline{0.133} & \uline{0.133} & \uline{0.133} & 0.134 & \uline{0.133} \\ 
        \multicolumn{1}{c|}{} & 192 & 0.21 & \textbf{0.148} & 0.181 & \uline{0.15} & \uline{0.15} & \uline{0.15} & 0.151 & \uline{0.15} & \uline{0.15} \\ 
        \multicolumn{1}{c|}{} & 336 & 0.224 & 0.165 & 0.196 & 0.162 & \textbf{0.159} & \uline{0.161} & \uline{0.161} & 0.163 & \textbf{0.159} \\ 
        \multicolumn{1}{c|}{} & 720 & 0.251 & 0.204 & 0.224 & \uline{0.197} & \uline{0.197} & \textbf{0.194} & \textbf{0.194} & \textbf{0.194} & \textbf{0.194} \\ 
        \hline 
        \multirow{4}*{\rotatebox{90}{Traffic}} &96& 0.618 & \textbf{0.358} & 0.468 & 0.382 & 0.384 & 0.384 & \uline{0.374} & 0.387 & \uline{0.374} \\ 
        \multicolumn{1}{c|}{} & 192 & 0.624 & \textbf{0.379} & 0.483 & 0.401 & 0.403 & 0.401 & 0.404 & \uline{0.4} & \uline{0.4} \\ 
        \multicolumn{1}{c|}{} & 336 & 0.63 & \textbf{0.388} & 0.493 & \uline{0.411} & 0.412 & 0.413 & 0.413 & 0.412 & \uline{0.411} \\ 
        \multicolumn{1}{c|}{} & 720 & 0.66 & \textbf{0.426} & 0.524 & 0.448 & 0.457 & \uline{0.445} & 0.454 & 0.451 & \uline{0.445} \\ 
        \hline 
        \multirow{4}*{\rotatebox{90}{Weather}} &96& 0.159 & 0.15 & 0.15 & 0.149 & \textbf{0.146} & 0.148 & \uline{0.147} & 0.15 & \textbf{0.146} \\ 
        \multicolumn{1}{c|}{} & 192 & 0.207 & \uline{0.195} & 0.196 & 0.196 & 0.196 & \textbf{0.194} & 0.197 & 0.197 & \textbf{0.194} \\ 
        \multicolumn{1}{c|}{} & 336 & 0.256 & \textbf{0.246} & \uline{0.248} & \textbf{0.246} & 0.251 & 0.249 & 0.249 & \textbf{0.246} & \textbf{0.246} \\ 
        \multicolumn{1}{c|}{} & 720 & 0.33 & 0.323 & 0.338 & \textbf{0.317} & \uline{0.318} & 0.319 & 0.321 & 0.319 & \textbf{0.317} \\ 
        \hline 
        \multicolumn{3}{c|}{\textbf{\% improvement over \vtsm}}& \textbf{18\% }&  \textbf{13.2\%} & \multicolumn{5}{c}{} & \textbf{19\%} \\ 
        \hline 
    \end{tabular}
    }
    \caption{Detailed MSE Analysis of various Channel Mixing techniques with different context lengths. Context length is varied from (1) to (5) and the minimum is selected as (Best) in this table. From the complete data analysis, we observe that ~\citsm~ outperforms~\vtsm~ by 18\%, and by adding a cross-channel reconciliation head (CC), the accuracy further improves by 1\% leading to a total improvement of 19\%. In contrast - ~\ictsm~ outperforms ~\vtsm~ but not ~\citsm~.}
    \label{tab:cc-detail}
\end{table*}

        

\begin{table*}
\centering
\captionsetup{width=.8\linewidth}
\setlength{\tabcolsep}{2pt}
\scalebox{0.9}{
\begin{tabular}{c|c|c|c|c|c|c}
\hline  
& \makecell{FL} & \makecell{CI-\tsm\\-Best} & \makecell{CI-\tsm\\(\ga,\hr))} & \makecell{CI-\tsm\\(\ga,\hr,\cc))} & \makecell{PatchTST} & \makecell{BTSF}  \\
\hline
\multirow{5}*{\rotatebox{90}{ETTH1}} & 24 & \textbf{0.314} & \uline{0.319} & \textbf{0.314} & 0.322 & 0.541 \\ 
\multicolumn{1}{c|}{} & 48 & \textbf{0.343} & \uline{0.344} & \textbf{0.343} & 0.354 & 0.613 \\ 
\multicolumn{1}{c|}{} & 168 & \textbf{0.397} & \textbf{0.397} & \uline{0.402} & 0.419 & 0.64 \\ 
\multicolumn{1}{c|}{} & 336 & \textbf{0.424} & \textbf{0.424} & \uline{0.43} & 0.445 & 0.864 \\ 
\multicolumn{1}{c|}{} & 720 & \textbf{0.453} & \textbf{0.453} & \uline{0.457} & 0.487 & 0.993 \\ 
\hline
\multirow{5}*{\rotatebox{90}{Weather}} & 24 & \uline{0.088} & 0.09 & \uline{0.088} & \textbf{0.087} & 0.324 \\ 
\multicolumn{1}{c|}{} & 48 & \uline{0.114} & \uline{0.114} & 0.141 & \textbf{0.113} & 0.366 \\ 
\multicolumn{1}{c|}{} & 168 & \textbf{0.177} & \textbf{0.177} & \uline{0.178} & \uline{0.178} & 0.543 \\ 
\multicolumn{1}{c|}{} & 336 & \textbf{0.241} & \textbf{0.241} & \uline{0.244} & \uline{0.244} & 0.568 \\ 
\multicolumn{1}{c|}{} & 720 & \textbf{0.319} & \textbf{0.319} & 0.322 & \uline{0.321} & 0.601 \\ 
\hline
\end{tabular}}
\caption{MSE Drill-down view of ~\tsmb~ for self-supervised multivariate forecasting. Min of ~\tsmgh~and~\tsmghc~is depicted as ~\tsmb, wherein. based on the considered dataset - either ~\tsmgh~ or ~\tsmghc~ outperforms the existing benchmarks.}
\label{tab:ab_fm_1}
\end{table*}

\end{document}

%% file: math_commands.tex
\usepackage{amsmath,amsfonts,bm}

\def\eqref#1{equation~\ref{#1}}

\def\1{\bm{1}}

\def\mH{{\bm{H}}}

\def\mW{{\bm{W}}}
\def\mX{{\bm{X}}}
\def\mY{{\bm{Y}}}

\DeclareMathAlphabet{\mathsfit}{\encodingdefault}{\sfdefault}{m}{sl}
\SetMathAlphabet{\mathsfit}{bold}{\encodingdefault}{\sfdefault}{bx}{n}

\def\gA{{\mathcal{A}}}

\def\gL{{\mathcal{L}}}
\def\gM{{\mathcal{M}}}

